\documentclass[sigconf]{acmart}

\usepackage[utf8]{inputenc} 
\usepackage[T1]{fontenc}    
\usepackage{hyperref}       
\usepackage{url}            
\usepackage{booktabs}       
\usepackage{amsfonts}       
\usepackage{nicefrac}       
\usepackage{microtype}      
\usepackage{algorithm}
\usepackage{xspace}
\usepackage{multirow}
\usepackage{dsfont}
\usepackage{amsmath} 
\usepackage{graphicx}
\usepackage{subcaption}
\usepackage[noend]{algpseudocode}
\usepackage{longtable}

\def\vec#1{\mathchoice%
	{\mbox{\boldmath $\displaystyle\bf#1$}}
	{\mbox{\boldmath $\textstyle\bf#1$}}
	{\mbox{\boldmath $\scriptstyle\bf#1$}}
	{\mbox{\boldmath $\scriptscriptstyle\bf#1$}}}
\def\v#1{\protect\vec #1}

\newcommand{\eg}{{\it e.g.\xspace}}
\newcommand{\etc}{{\it etc\xspace}}

\newcommand{\TODOK}[1]{
\ifmmode
\text{\textcolor{blue}{ }}
\else
\textcolor{blue}{ }
\fi
}
\newcommand{\TODOA}[1]{
\ifmmode
\text{\textcolor{red}{ }}
\else
\textcolor{red}{ }
\fi
}

\renewcommand{\vec}[1]{{\mathbf{#1}}}
\newcommand{\br}[1]{\left({#1}\right)}

\makeatletter

\usepackage{amsthm}

\makeatletter
\newtheorem*{rep@theorem}{\rep@title}
\newcommand{\newreptheorem}[2]{%
\newenvironment{rep#1}[1]{%
 \def\rep@title{#2 \ref{##1}}%
 \begin{rep@theorem}}%
 {\end{rep@theorem}}}
\makeatother

\newtheorem{theorem}{Theorem}
\newreptheorem{theorem}{Theorem}

\newreptheorem{lemma}{Lemma}

\numberwithin{theorem}{section}

\newcommand{\bigO}[1]{{\mathcal O}\br{{#1}}}

\newcommand{\Om}[1]{\Omega\br{{#1}}}

\makeatletter
\newcommand{\opnorm}{\@ifstar\@opnorms\@opnorm}

\newcommand{\argmin}[1]{\underset{#1}{\operatorname{arg}\,\operatorname{min}}\;}

\usepackage{array}

\newcolumntype{L}{>{\arraybackslash}m{8cm}}

\newcommand{\vx}{\mathbf{x}_{i}}

\definecolor{arsenic}{rgb}{0.23, 0.27, 0.29}

\definecolor{silver}{rgb}{0.75, 0.75, 0.75}
\newcommand{\wpred}[1]{
\textcolor{silver}{#1}
}

\newcommand{\suppl}{\href{http://manikvarma.org/pubs/mittal21.pdf}{\color{blue}{supplementary material}}\xspace}

\newcommand{\code}{\href{https://github.com/Extreme-classification/DECAF}{\color{blue}{https://github.com/Extreme-classification/DECAF}}}\xspace

\newcommand{\vc}{\v c}
\newcommand{\ve}{\v e}

\newcommand{\vh}{\v h}
\newcommand{\vr}{\v r}
\newcommand{\vu}{\v u}

\newcommand{\vw}{\v w}
\renewcommand{\vx}{\v x}
\newcommand{\vy}{\v y}
\newcommand{\vz}{\v z}
\newcommand{\vE}{\v E}
\newcommand{\vR}{\v R}
\newcommand{\vH}{\v H}
\newcommand{\vW}{\v W}
\newcommand{\vzero}{\v0}
\newcommand{\vone}{\v1}

\newcommand{\hvz}{\hat\vz}

\def\mydefgreek#1{\expandafter\def\csname v#1\endcsname{\text{\boldmath$\mathbf{\csname #1\endcsname}$}}}
\def\mydefallgreek#1{\ifx\mydefallgreek#1\else\mydefgreek{#1}%
   \lowercase{\mydefgreek{#1}}\expandafter\mydefallgreek\fi}
\mydefallgreek {alpha}{beta}{gamma}{delta}{epsilon}{zeta}{eta}{theta}{iota}{kappa}{lambda}{mu}{nu}{xi}{omicron}{pi}{rho}{sigma}{tau}{upsilon}{phi}{chi}{psi}{omega}\mydefallgreek

\def\LatinUpper{A,B,C,D,E,F,G,H,I,J,K,L,M,N,O,P,Q,R,S,T,U,V,W,X,Y,Z}

\newcommand{\genCal}[1]{\expandafter\newcommand\csname c#1\endcsname{{\mathcal #1}}}
\@for\i:=\LatinUpper\do{%
	\expandafter\genCal\i
}

\newcommand{\genBb}[1]{\expandafter\newcommand\csname b#1\endcsname{{\mathbb #1}}}
\@for\i:=\LatinUpper\do{%
	\expandafter\genBb\i
}

\newcommand{\bc}[1]{\left\{{#1}\right\}}
\renewcommand{\br}[1]{\left({#1}\right)}
\newcommand{\bs}[1]{\left[{#1}\right]}

\newcommand{\norm}[1]{\left\| {#1} \right\|}

\newcommand{\ip}[2]{\left\langle{#1},{#2}\right\rangle}

\renewcommand{\P}[1]{\bP\bs{{#1}}}

\renewcommand{\argmin}{\mathop{\arg\min}}

\newcommand{\nth}{\text{$^{\text{th}}$}\xspace}

\newcommand{\cond}{\,|\,}
\usepackage{cancel}

\copyrightyear{2021}
\acmYear{2021}
\setcopyright{acmlicensed}
\acmConference[WSDM '21]{Proceedings of the Fourteenth ACM International Conference on Web Search and Data Mining}{March 8--12, 2021}{Virtual Event, Israel}
\acmBooktitle{Proceedings of the Fourteenth ACM International Conference on Web Search and Data Mining (WSDM '21), March 8--12, 2021, Virtual Event, Israel}
\acmPrice{15.00}
\acmDOI{10.1145/3437963.3441807}
\acmISBN{978-1-4503-8297-7/21/03}

\def\algfull{Deep Extreme Classification with Label Features}
\def\alg{\textsf{DECAF}\xspace}

\def\algl{\textsf{DECAF-lite}\xspace}

\begin{document}
\fancyhead{}

\title{\alg: \algfull}

\author{Anshul Mittal}
\email{me@anshulmittal.org}
\author{Kunal Dahiya}
\email{kunalsdahiya@gmail.com}
\affiliation{%
    \institution{IIT Delhi}
    \country{India}
}
\author{Sheshansh Agrawal}
\email{sheshansh.agrawal@microsoft.com}
\author{Deepak Saini}
\email{desaini@microsoft.com}
\affiliation{%
    \institution{Microsoft Research}
    \country{India}
}
\author{Sumeet Agarwal}
\orcid{0000-0002-5714-3921}
\email{sumeet@iitd.ac.in}
\affiliation{%
    \institution{IIT Delhi}
    \country{India}
}
\author{Purushottam Kar}
\orcid{0000-0003-2096-5267}
\email{purushot@cse.iitk.ac.in}
\affiliation{%
    \institution{IIT Kanpur}
    \institution{Microsoft Research}
    \country{India}
}
\author{Manik Varma}
\email{manik@microsoft.com}
\affiliation{%
    \institution{Microsoft Research}
    \institution{IIT Delhi}
    \country{India}
}


\renewcommand{\shortauthors}{Mittal and Dahiya, et al.}

\begin{abstract}
Extreme multi-label classification (XML) involves tagging a data point with its most relevant subset of labels from an extremely large label set, with several applications such as product-to-product recommendation with millions of products. Although leading XML algorithms scale to millions of labels, they largely ignore label metadata such as textual descriptions of the labels. On the other hand, classical techniques that can utilize label metadata via representation learning using deep networks struggle in extreme settings. This paper develops the \alg algorithm that addresses these challenges by learning models enriched by label metadata that jointly learn model parameters and feature representations using deep networks and offer accurate classification at the scale of millions of labels. \alg makes specific contributions to model architecture design, initialization, and training, enabling it to offer up to 2-6$\%$ more accurate prediction than leading extreme classifiers on publicly available benchmark product-to-product recommendation datasets, such as LF-AmazonTitles-1.3M. At the same time, \alg was found to be up to 22$\times$ faster at inference than leading deep extreme classifiers, which makes it suitable for real-time applications that require predictions within a few milliseconds. The code for \alg is available at the following URL \newline \code.
\end{abstract}



\maketitle

\section{Introduction}
\label{sec:intro}

\textbf{Objective}: Extreme multi-label classification (XML) refers to the task of tagging data points with a relevant subset of labels from an extremely large label set. This paper demonstrates that XML algorithms stand to gain significantly by incorporating label metadata. The \alg algorithm is proposed, which could be up to 2-6\% more accurate than leading XML methods such as Astec~\citep{Dahiya21}, MACH~\cite{Medini2019}, Bonsai~\citep{Khandagale19}, AttentionXML~\citep{You18}, \etc, while offering predictions within a fraction of a millisecond, which makes it suitable for high-volume and time-critical applications.

\textbf{Short-text applications}: Applications such as predicting related products given a retail product's name \cite{Medini2019}, or predicting related webpages given a webpage title, or related searches \cite{Jain19}, all involve short texts, with the product name, webpage title, or search query having just 3-10 words on average. In addition to the statistical and computational challenges posed by a large set of labels, short-text tasks are particularly challenging as only a few words are available per data point. This paper focuses on short-text applications such as related product and webpage recommendation.

\textbf{Label metadata}: Metadata for labels can be available in various forms: textual representations, label hierarchies, label taxonomies \cite{Kanagal12, menon11, sachdeva19}, or label correlation graphs, and can capture semantic relations between labels. For instance, the Amazon products (that serve as labels in a product-to-product recommendation task) ``Panzer Dragoon", and ``Panzer Dragoon Orta" do not share any common training point but are semantically related. Label metadata can allow collaborative learning, which especially benefits \emph{tail} labels. Tail labels are those for which very few training points are available and form the majority of labels in XML applications~\citep{Jain16, Babbar17, Babbar19}. For instance, just 14 documents are tagged with the label ``Panzer Dragoon Orta" while 23 documents are tagged with the label ``Panzer Dragoon" in the LF-AmazonTitles-131K dataset. In this paper, we will focus on utilizing label text as a form of label metadata.

\textbf{\alg}: \alg learns a separate linear classifier per label based on the 1-vs-All approach. These classifiers critically utilize label metadata and require careful initialization since random initialization~\citep{Glorot2010} leads to inferior performance at extreme scales. \alg proposes using a \emph{shortlister} with large fanout to cut down training and prediction time drastically. Specifically, given a training set of $N$ examples, $L$ labels, and $D$ dimensional embeddings being learnt, the use of the shortlister brings training time down from $\bigO{NDL}$ to $\bigO{ND\log L}$ (by training only on the $\bigO{\log L}$ most confusing negative labels for every training point), and prediction time down from $\bigO{DL}$ to $\bigO{D\log L}$ (by evaluating classifiers corresponding to only the $\bigO{\log L}$ most likely labels). An efficient and scalable two-stage strategy is proposed to train the shortlister.

\textbf{Comparison with state-of-the-art}: Experiments conducted on publicly available benchmark datasets revealed that \alg could be 5\% more accurate than the leading approaches such as DiSMEC~\citep{Babbar17}, Parabel~\citep{Prabhu18b}, Bonsai~\cite{Khandagale19} AnnexML~\citep{Tagami17}, \etc, which utilize pre-computed features. \alg was also found to be 2-6\% more accurate than leading deep learning-based approaches such as Astec~\citep{Dahiya21}, AttentionXML~\citep{You18} and MACH~\citep{Medini2019} that jointly learn feature representations and classifiers. Furthermore, \alg could be up to 22$\times$ faster at prediction than deep learning methods such as MACH and AttentionXML.

\textbf{Contributions}: This paper presents \alg, a scalable deep learning architecture for XML applications that effectively utilize label metadata. Specific contributions are made in designing a shortlister with a large fanout and a two-stage training strategy. \alg also introduces a novel initialization strategy for classifiers that leads to accuracy gains, more prominently on data-scarce tail labels. \alg scales to XML tasks with millions of labels and makes predictions significantly more accurate than state-of-the-art XML methods. Even on datasets with more than a million labels, \alg can make predictions in a fraction of a millisecond, thereby making it suitable for real-time applications.

\section{Related Work}
\label{sec:related}

\textbf{Summary}: XML techniques can be categorized into 1-vs-All, tree, and embedding methods. Of these, one-vs-all methods such as Slice \cite{Jain19} and Parabel \cite{Prabhu18b} offer the most accurate solutions. Recent advances have introduced the use of deep-learning-based representations. However, these techniques mostly do not use label metadata. Techniques such as the X-Transformer \cite{Chang20} that do use label text either do not scale well with millions of labels or else do not offer state-of-the-art accuracies. The \alg method presented in this paper effectively uses label metadata to offer state-of-the-art accuracies and scale to tasks with millions of labels.

\textbf{1-vs-All classifiers}: 1-vs-All classifiers PPDSparse~\citep{Yen17}, DiSMEC~\cite{Babbar17}, ProXML~\cite{Babbar19} are known to offer accurate predictions but risk incurring training and prediction costs that are linear in the number of labels, which is prohibitive at extreme scales. Approaches such as negative sampling, PLTs, and learned label hierarchies have been proposed to speed up training~\citep{Jain19, Khandagale19, Prabhu18b, Yen18a}, and predictions~\citep{Jasinska16, Niculescu17} for 1-vs-All methods. However, they rely on sub-linear search structures such as nearest-neighbor structures or label-trees that are well suited for fixed or pre-trained features such as bag-of-words or FastText~\citep{Joulin17} but not support jointly learning deep representations since it is expensive to repeatedly update these search structures as deep-learned representations keep getting updated across learning epochs. Thus, these approaches are unable to utilize deep-learned features, which leads to inaccurate solutions. \alg avoids these issues by its use of the \emph{shortlister} which offers a high recall filtering of labels allowing training and prediction costs that are logarithmic in the number of labels.

\textbf{Tree classifiers}: Tree-based classifiers typically partition the label space to achieve logarithmic prediction complexity. In particular, MLRF~\citep{Agrawal13}, FastXML~\citep{Prabhu14}, PfastreXML~\citep{Jain16} learn an ensemble of trees where each node in a tree is partitioned by optimizing an objective based on the Gini index or nDCG. CRAFTML~\cite{Siblini18a} deploys random partitioning of features and labels to learn an ensemble of trees. However, such algorithms can be expensive in terms of training time and model size.

\textbf{Deep feature representations}: Recent works MACH~\citep{Medini2019}, X-Transformer \cite{Chang20}, XML-CNN~\citep{Liu17}, and AttentionXML~\citep{Liu17} have graduated from using fixed or pre-learned features to using task-specific feature representations that can be significantly more accurate. However, CNN and attention-based mechanisms were found to be inaccurate on short-text applications (as shown in \cite{Dahiya21}) where scant information is available (3-10 tokens) for a data point. Furthermore, approaches like X-Transformer and AttentionXML that learn label-specific document representations do not scale well.

\textbf{Using label metadata}: Techniques that use label metadata e.g. label text include SwiftXML \citep{Prabhu18} which uses a pre-trained Word2Vec \citep{Mikolov13} model to compute label representations. However, SwiftXML is designed for \emph{warm-start} settings where a subset of ground-truth labels for each test point is already available. This is a non-standard scenario that is beyond the scope of this paper. \citep{Guo2019} demonstrated, using the GlaS regularizer, that modeling label correlations could lead to gains on tail labels. Siamese networks~\citep{Wu17} are a popular framework that can learn representations so that documents and their associated labels get embedded together. Unfortunately, Siamese networks were found to be inaccurate at extreme scales. The X-Transformer method \cite{Chang20} uses label text to generate shortlists to speed up training and prediction. \alg, on the other hand, makes much more direct use of label text to train the 1-vs-All label classifiers themselves and offers greater accuracy compared to X-Transformer and other XML techniques that also use label text.

\section{\alg: \algfull}
\label{sec:method}

\begin{figure}
    \centering
    \includegraphics[width=\columnwidth]{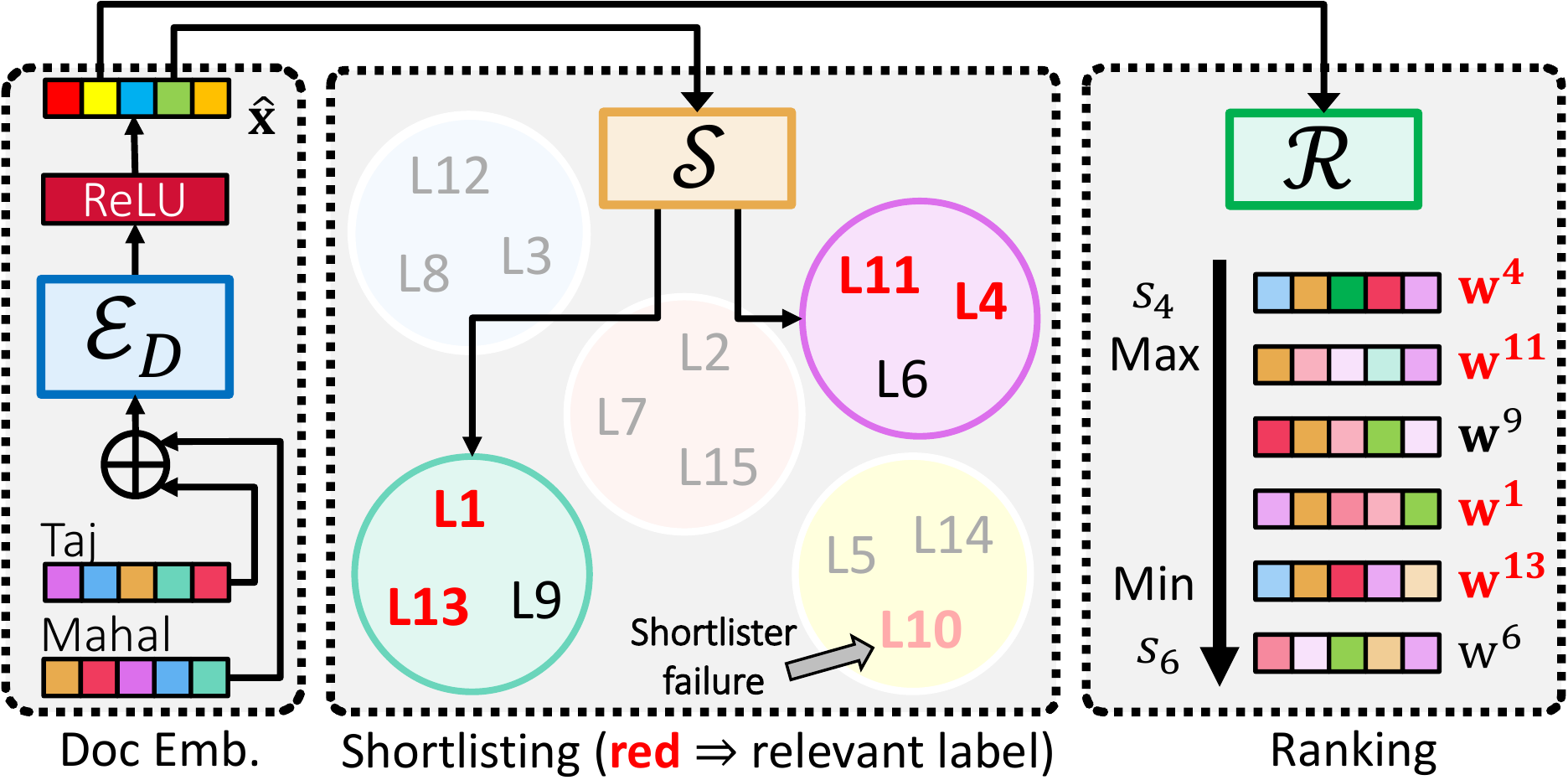}
    \caption{\alg's frugal prediction pipeline scales to millions of labels. Given a document $\vx$, its text embedding $\hat\vx$ (see Fig~\ref{fig:embedding} (Left)) is first used by the shortlister $\cS$ to shortlist the most probable $\bigO{\log L}$ labels while maintaining high recall. The ranker $\cR$ then uses label classifiers (see Fig~\ref{fig:embedding} (Right)) of only the shortlisted labels to produce the final ranking.}
    \label{fig:architecture}
\end{figure}

\begin{figure}
    \centering
    \includegraphics[width=0.91\columnwidth]{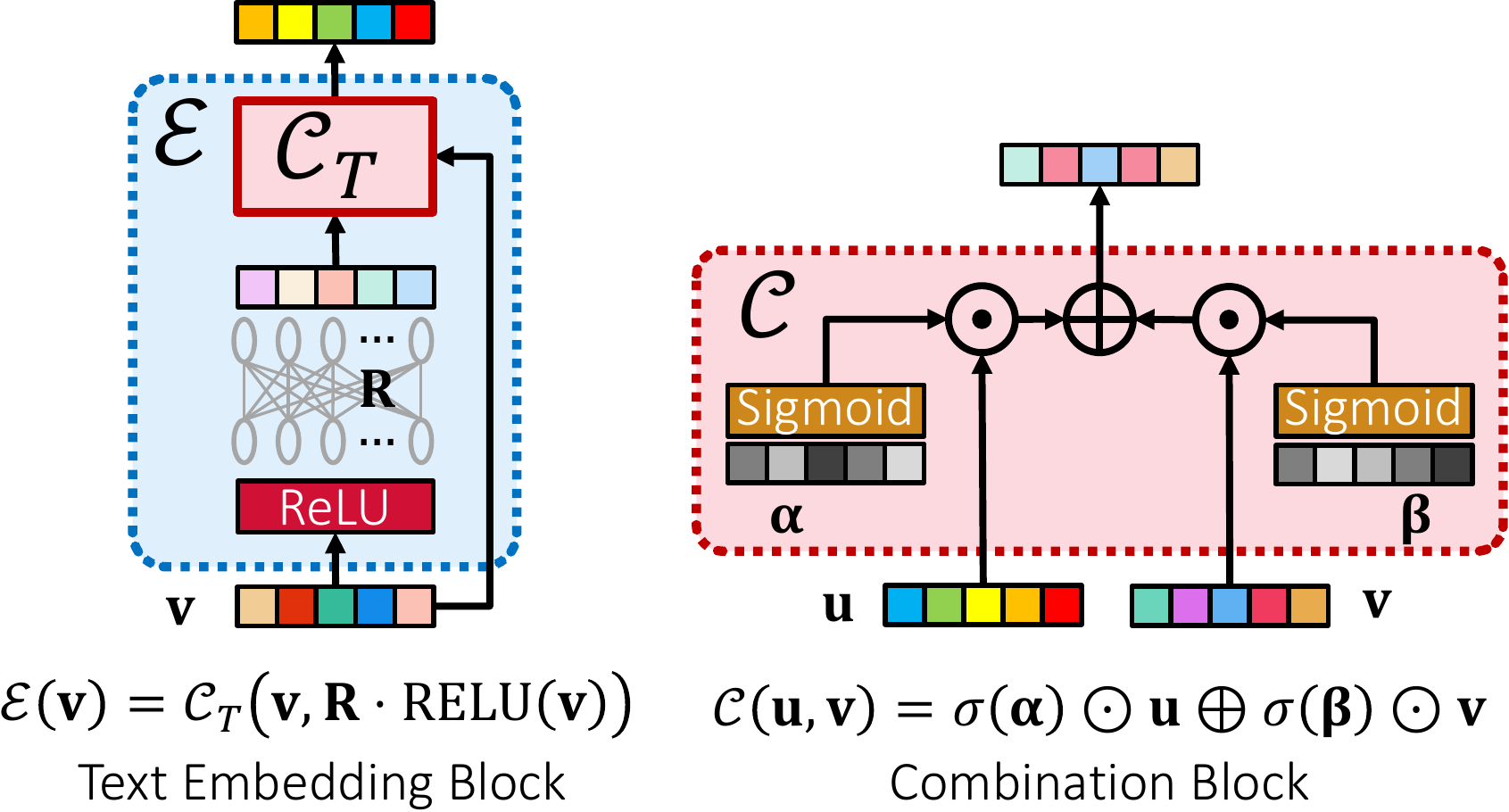}
    \caption{(Left) \alg uses a lightweight architecture with a residual layer to embed both document and label text (see Fig.~\ref{fig:embedding}). (Right) Combination blocks are used to combine various representations (separate instances are used in the text embedding blocks ($\cE_D, \cE_L$) and in label classifiers ($\cC_L$)).}
    \label{fig:blocks}
\end{figure}

\begin{figure}
    \centering
    \includegraphics[width=\columnwidth]{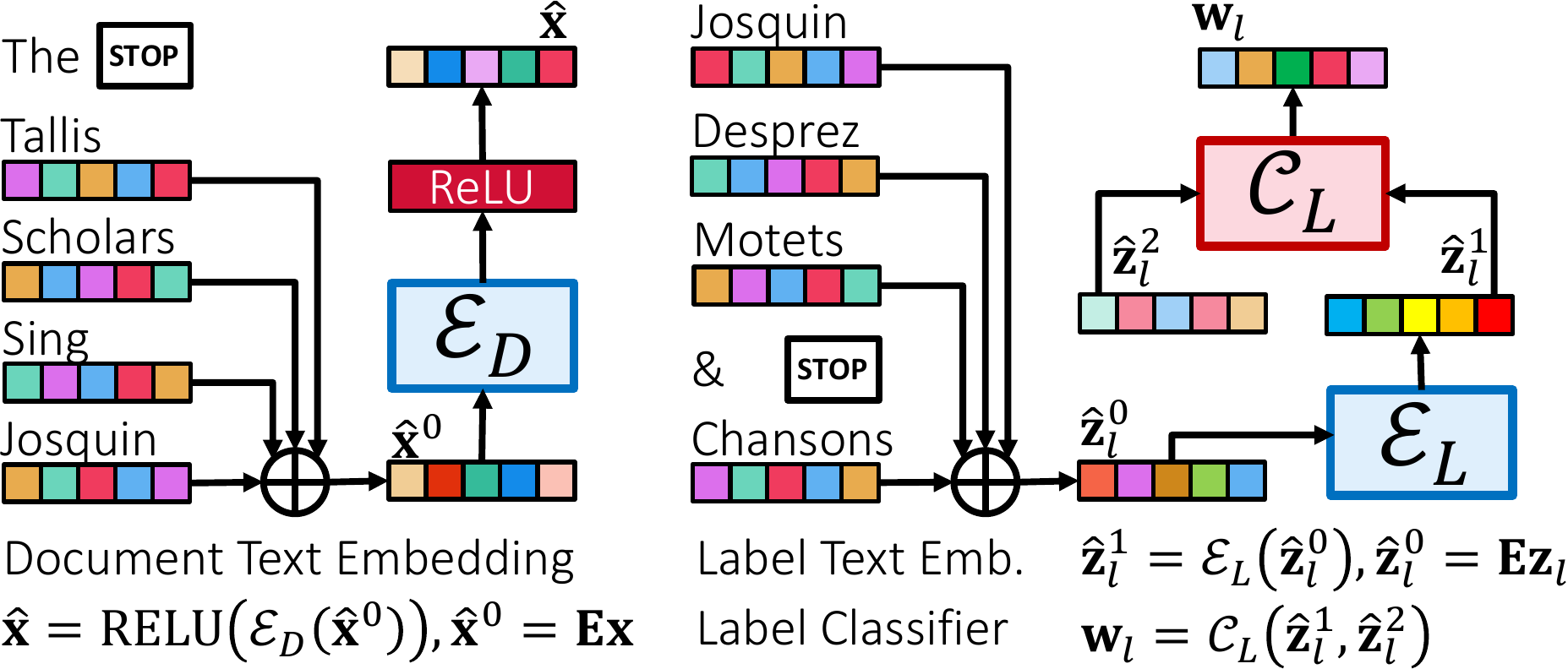}
    \caption{(Left) Document text is embedded using an instance $\cE_D$ of the text embedding block (see Fig.~\ref{fig:blocks}). Stop words (e.g. \emph{and, the}) are discarded. (Right) \alg critically incorporates label text into classifier learning. For each label $l \in [L]$, a one-vs-all classifier $\vw_l$ is learnt by combining label text embedding $\hvz^1_l$ (using a separate instance $\cE_L$ of the text embedding block) and a refinement vector $\hvz^2_l$. Note that $\cE_D,\cE_L,\cC_L$ use separate parameters. However, all labels share the blocks $\cE_L, \cC_L$ and all documents share the block $\cE_D$.}
    \label{fig:embedding}
\end{figure}

\textbf{Summary}: \alg consists of three components 1) a lightweight text embedding block suitable for short-text applications, 2) 1-vs-All classifiers per label that incorporate label text, and 3) a shortlister that offers a high recall label shortlists for data points, allowing \alg to offer sub-millisecond prediction times even with millions of labels. This section details these components, and an approximate likelihood model with provable recovery guarantees, using which \alg offers a highly scalable yet accurate pipeline for jointly training text embeddings and classifier parameters.


\textbf{Notation}: Let $L$ be the number of labels and $V$ be the dictionary size. Each of the $N$ training points is presented as $(\vx_i,\vy_i)$. $\vx_i \in \bR^V$ is a bag-of-tokens representation for the $i\nth$ document i.e. $x_{it}$ is the TF-IDF weight of token $t \in [V]$ in the $i\nth$ document. $\vy_i \in \bc{-1,+1}^L$ is the ground truth label vector with $y_{il} = +1$ if label $l \in [L]$ is relevant to the $i\nth$ document and $y_{il} = -1$ otherwise. For each label $l \in [L]$, its label text is similarly represented as $\vz_l \in \bR^V$.

\textbf{Document and label-text embedding}: \alg learns $D$-dim embeddings for each vocabulary token i.e. $\vE = \bs{\ve_1,\ldots,\ve_V} \in \bR^{D \times V}$ and uses a light-weight embedding block (see Fig~\ref{fig:embedding}) to encode label and document texts. The embedding block $\cE = \bc{\vR,\valpha,\vbeta}$ is parameterized by a residual block $\vR \in \bR^{d \times d}$ and scaling constants $\valpha,\vbeta \in \bR^D$ for the combination block (see Fig~\ref{fig:blocks}). The embedding for a bag-of-tokens vector, say $\vr \in \bR^V$, is $\cE(\vr) = \sigma(\valpha) \odot \hat\vr^0 + \sigma(\vbeta) \odot (\vR\cdot \text{ReLU}(\hat\vr^0)) \in \bR^D$ where $\hat\vr^0 = \vE\vr$, $\odot$ denotes component-wise multiplication, and $\sigma$ is the sigmoid function. Document embeddings, denoted by $\hat\vx_i$, are computed as $\hat\vx_i = \text{ReLU}(\cE_D(\vx_i))$. Label-text embeddings, denoted by $\hat\vz^1_l$ are computed as $\hat\vz^1_l = \cE_L(\vz_l)$. Note that document and labels use separate instantiations $\cE_D, \cE_L$ of the embedding block. We note that \alg could also be made to use alternate text representations such as BERT~\cite{Devlin19}, attention~\cite{You18}, LSTM~\citep{hochreiter97} or convolution~\cite{Liu17}. However, such elaborate architectures negatively impact prediction time and moreover, \alg outperforms BERT, CNN and attention based XML techniques on all our benchmark datasets indicating the suitability of \alg's frugal architecture to short-text applications. 

\textbf{1-vs-All Label Classifiers}: \alg uses high capacity 1-vs-All (OvA) classifiers $\vW = [\vw_1,\ldots,\vw_L] \in \bR^{D \times L}$ that outperform tree- and embedding-based classifiers \cite{Chang20,Jain19,Babbar19,Prabhu18b,Yen17,Babbar17}. However, \alg distinguishes itself from previous OvA works (even those such as \cite{Chang20} that do use label text) by directly incorporating label text into the OvA classifiers. For each label $l \in [L]$, the label-text embedding $\hat\vz^1_l = \cE_L(\vz_l)$ (see above) is combined with a \emph{refinement} vector $\hat\vz^2_l$ that is learnt separately per label, to produce the label classifier $\vw_l = \sigma(\valpha_L) \odot \hat\vz^1_l + \sigma(\vbeta_L) \odot \hat\vz^2_l \in \bR^D$ where $\valpha_L, \vbeta_L \in \bR^D$ are shared across labels (see Fig~\ref{fig:embedding}). Incorporating $\hat\vz^1_l$ into the label classifier $\vw_l$ allows labels that never co-occur, but nevertheless share tokens, to perform learning in a collaborative manner since if two labels, say $l,m \in [L]$ share some token $t \in [V]$ in their respective texts, then $\ve_t$ contributes to both $\hat\vz^1_l$ and $\hat\vz^1_m$. In particular, this allows rare labels to share classifier information with popular labels with which they share a token. Ablation studies (Tab~\ref{tab:bowmeta},\ref{tab:sub:xmlclass},\ref{tab:combouv}) show that incorporating label text into classifier learning offers \alg significant gains of over 2-6\% compared to methods that do not use label text. Incorporating other forms of label metadata, such as label hierarchies, could also lead to further gains.

\textbf{Shortlister}: OvA training and prediction can be prohibitive, $\Om{NDL}$ and $\Om{DL}$ resp., if done naively. A popular way to accelerate training is to, for every data point $i \in [N]$, use only a \emph{shortlist} containing all positive labels (that are relatively fewer around $\bigO{\log L}$) and a small subset of the, say again $\bigO{\log L}$, most challenging negative labels \cite{Chang20,Jain19,Khandagale19,Prabhu18b,Yen17,Bhatia15}. This allows training to be performed in $\bigO{ND\log L}$ time instead of $\bigO{NDL}$ time. \alg learns a \emph{shortlister} $\cS$ that offers a label-clustering based shortlisting. We have $\cS = \bc{\cC, \vH}$ where $\cC = \bc{C_1,\ldots,C_K}$ is a balanced clustering of the $L$ labels and $\vH = [\vh_1,\ldots,\vh_K] \in \bR^{D \times K}$ are OvA classifiers, one for each cluster. Given the embedding $\hat\vx$ of a document and \emph{beam-size} $B$, the top $B$ clusters with the highest scores, say $\ip{\vh_{m_1}}{\hat\vx} \geq \ip{\vh_{m_2}}{\hat\vx} \geq \ldots$ are taken and labels present therein are shortlisted i.e. $\cS(\hat\vx) := \bc{m_1,\ldots,m_B}$. As clusters are balanced, we get, for every datapoint, $LB/K$ shortlisted labels in the clusters returned. \alg uses $K = 2^{17}$ clusters for large datasets.

\textbf{Prediction Pipeline}: Fig~\ref{fig:architecture} shows the frugal prediction pipeline adopted by \alg. Given a document $\vx \in \bR^V$, its embedding $\hat\vx = \text{ReLU}(\cE_D(\vx))$ is used by the shortlister to obtain a shortlist of $B$ label clusters $\cS(\hat\vx) = \bc{m_1,\ldots,m_B}$. Label scores are computed for every shortlisted label i.e. $l \in C_m, m \in \cS(\hat\vx)$ by combining shortlister and OvA classifier scores as $s_l := \sigma(\ip{\vw_l}{\hat\vx})\cdot\sigma(\ip{\vh_m}{\hat\vx})$. These scores are sorted to make the final prediction. In practice, even on a dataset with 1.3 million labels, \alg could make predictions within 0.2 ms using a GPU and 2 ms using a CPU.

\subsection{Efficient Training: the DeepXML Pipeline}
\textbf{Summary}: \alg adopts the scalable DeepXML pipeline \cite{Dahiya21} that splits training into 4 \emph{modules}. In summary, Module I jointly learns the token embeddings $\vE$, the embedding modules $\cE_D, \cE_L$ and shortlister $\cS$. Module II fine-tunes $\cE_D, \cE_L, \cS$, and retrieves label shortlists for all data points. After performing initialization in Module III, Module IV jointly learns the OvA classifiers $\vW$ and fine-tunes $\cE_D, \cE_L$ using the shortlists generated in Module II. Due to lack of space some details are provided in the \suppl\footnote{Supplementary Material Link: \color{blue}{\url{http://manikvarma.org/pubs/mittal21.pdf}}}

\textbf{Module I}: Token embeddings $\vE \in \bR^{D \times V}$ are randomly initialized using~\cite{he2015delving}, residual blocks within the blocks $\cE_D, \cE_L$ are initialized to identity, and label \emph{centroids} are created by aggregating document information for each label $l \in [L]$ as $\vc_l = \sum_{i:y_{il} = +1}\vx_i$. Balanced hierarchical binary clustering \cite{Prabhu18b} is now done on these label centroids for 17 levels to generate $K$ label clusters. Clustering labels using label centroids gave superior performance than using other representations such as label text $\vz_l$. This is because the label centroid carries information from multiple documents and thus, a diverse set of tokens whereas $\vz_l$ contains information from only a handful of tokens. The hierarchy itself is discarded and each resulting cluster is now treated as a \emph{meta-label} that gives us a \emph{meta} multi-label classification problem on the same training points, but with $K$ meta-labels instead of the original $L$ labels. Each meta label $m \in [K]$ is granted meta-label text as $\vu_m = \sum_{l \in C_m}\vz_l$. Each datapoint $i \in [N]$ is assigned a meta-label vector $\tilde\vy_i \in \bc{-1,+1}^K$ such that $\tilde y_{im} = +1$ if $y_{il} = +1$ for any $l \in C_m$ and $\tilde y_{im} = -1$ if $y_{il} = -1$ for all $l \in C_m$. OvA meta-classifiers $\vH = [\vh_1,\ldots,\vh_K] \in \bR^{D \times K}$ are learnt to solve this meta multi-label problem but are constrained in Module I to be of the form $\vh_m = \cE_L(\vu_m)$. This constrained form of the meta-classifier forces good token embeddings $\vE$ to be learnt that allow meta-classification without the assistance of powerful refinement vectors. However, this form continues to allow collaborative learning among meta classifiers based on shared tokens. Module I solves the meta multi-label classification problem while jointly training $\cE_D, \cE_L, \vE$ (implicitly learning $\vH$ in the process).

\textbf{Module II}: The shortlister is fine-tuned in this module. Label centroids are recomputed as $\vc_l = \sum_{i:\vy^i_l = +1}\vE\vx_i$ where $\vE$ are the task-specific token embeddings learnt in Module I. The meta multi-label classification problem is recreated using these new centroids by following the same steps outlined in Module I. Module II uses OvA meta-classifiers that are more powerful and resemble those used by \alg. Specifically, we now have $\vh_m = \sigma(\tilde\valpha_P)\odot\hat\vu^2_m + \sigma(\tilde\vbeta_P)\odot\hat\vu^1_m$ where $\hat\vu^1_m = \sum_{l \in C_m}\cE_L(\vz_l)$ is the meta label-text embedding, $\hat\vu^2_m$ are meta label-specific refinement vectors, and $\cC_P = \bc{\tilde\valpha_P, \tilde\vbeta_P}$ is a fresh instantiating of the combination block. Module II solves the (new) meta multi-label classification problem, jointly learning $\cC_P, \hat\vu^2_m$ (implicitly updating $\vH$ in the process) and fine-tuning $\cE_D, \cE_L, \vE$. The shortlister $\cS$ so learnt is now used to retrieve shortlists $\cS(\vx_i)$ for each data point $i \in [N]$.

\textbf{Module III}: Residual blocks within $\cE_D, \cE_L$ are re-initialized to identity, $\cS$ is frozen and combination block parameters for the OvA classifiers are initialized to $\valpha_L = \vbeta_L = \vzero$ (note that $\sigma(\vzero) = 0.5\cdot\vone$ where $\vone$ is the all-ones vector). Refinement vectors for all $L$ labels are initialized to $\hat\vz^2_l = \vE\vz_l$. Ablation studies (see Tab~\ref{tab:combouv}) show that this refinement vector initialization offers performance boosts of up to 5-10\% compared to random initialization as is used by existing methods such as AttentionXML \cite{You18} and the X-Transformer \cite{Chang20}. 

\textbf{Module IV}: This module performs learning using an approximate likelihood model. Let $\Theta = \bc{\vE, \cE_D, \cE_L, \cC_L, \vW}$ be the model parameters in the \alg architecture. We recall that $\cC_L$ are combination blocks used to construct the OvA classifiers and meta classifiers, and $\vE$ are the token embeddings. OvA approaches assume a likelihood decomposition such as $\P{\vy_i \cond \vx_i, \Theta} = \prod_{l=1}^L\P{y_{il} \cond \hat\vx_i, \vw_l} = \prod_{l=1}^L\br{1 + \exp\br{-y_{il}\cdot\ip{\hat\vx_i}{\vw_l}}}^{-1}$. Here $\hat\vx_i = \text{ReLU}(\cE_D(\vx_i))$ is the document-text embedding and $\vw_l$ are the OvA classifiers as shown in Fig~\ref{fig:embedding}. Let us abbreviate $\ell_{il}(\Theta) = \ln\br{1 + \exp\br{-y_{il}\cdot\ip{\hat\vx_i}{\vw_l}}}$. Then, our objective is to optimize $\argmin_{\Theta} \cL(\Theta)$ where
\[
\cL(\Theta) = \frac1{NL}\sum_{i \in [N]}\sum_{l \in [L]}\ell_{il}(\Theta)
\]
However, performing the above optimization exactly is intractable and takes $\Om{NDL}$ time. \alg's solves this problem by instead optimizing $\argmin_\Theta \tilde\cL(\Theta \cond \cS)$ where
\[
\tilde\cL(\Theta \cond \cS) = \frac{K}{NLB}\sum_{i \in [N]}\sum_{l \in \cS(\hat\vx_i)}\ell_{il}(\Theta)
\]
Recall that for any document, $\cS(\hat\vx_i)$ is a shortlist of $B$ label clusters (that give us a total of $LB/K$ labels). Thus, the above expression contains only $NLB/K \ll NL$ terms as \alg uses a large fanout of $K \approx 130$K and $B \approx 100$. The result below assures us that model parameters and embeddings obtained by optimizing $\tilde\cL(\Theta \cond \cS)$ perform well w.r.t. the original likelihood $\cL(\Theta)$ if the dataset exhibits label sparsity, and the shortlister assures high recall. 

\begin{theorem}
\label{thm:thm}
Suppose the training data has label sparsity at rate $s$ i.e. $\sum_{i \in [N]}\sum_{l \in [L]} \bI\bc{y_{il} = +1} = s\cdot NL$ and the shortlister offers a recall rate of $r$ on the training set i.e. $\sum_{i \in [N]}\sum_{l \in \cS(\hat\vx_i)}\bI\bc{y_{il} = +1} = rs\cdot NL$. Then if $\hat\Theta$ is obtained by optimizing the approximate likelihood function $\tilde\cL(\Theta \cond \cS)$, then the following always holds
\[
\cL(\hat\Theta) \leq \min_{\Theta} \cL(\Theta) + \bigO{s(1-r)\ln(1/(s(1-r)))}.
\]
\end{theorem}
Please refer to Appendix~A.1 in the \suppl for the proof. As $s \rightarrow 0$ and $r \rightarrow 1$, the excess error term vanishes at rate at least $\sqrt{s(1-r)}$. Our XML datasets do exhibit label sparsity at rate $s \approx 10^{-5}$ and Fig~\ref{fig:recall_clusters} shows that \alg's shortlister does offer high recall with small shortlists (80\% recall with $\approx 50$-sized shortlist and 85\% recall with $\approx 100$-sized shortlist). Since Thm~\ref{thm:thm} holds in the completely agnostic setting, it establishes the utility of learning when likelihood maximization is performed only on label shortlists with high-recall. Module IV uses these shortlists to jointly learn the $L$ OvA classifiers $\vW$ and $\cC_L$, as well as fine-tune the embedding blocks $\cE_D, \cE_L$ and token embeddings $\vE$.


\textbf{Loss Function and Regularization}: Modules I, II, IV use the logistic loss and the Adam~\cite{Kingma14} optimizer to train the model parameters and various refinement vectors. Residual layers used in the text embedding blocks $\cE_D, \cE_L$ were subjected to spectral regularization \cite{Miyato18b}. All ReLU layers were followed by a dropout layer with 50\% drop-rate in Module-I and 20\% for the rest of the modules.


\textbf{Ensemble Learning}: \alg learns an inexpensive ensemble of 3 instances (see Figure~\ref{fig:numlearner}). The three instances share Module I training to promote scalability i.e. they inherit the same token embeddings. However, they carry out training Module II onwards independently. Thus, the shortlister and embedding modules get fine-tuned for each instance.


\textbf{Time Complexity}: Appendix~A.2 in the \suppl presents time complexity analysis for the \alg modules.
\section{Experiments}
\label{sec:results}

\begin{table}[ht]
    \caption{Results on publicly available short-text datasets. \alg was found to be 2--6\% more accurate, as well as an order of magnitude faster at prediction compared to other deep learning based approaches. Algorithms marked with a `-' were unable to scale on the given dataset within available resources and timeout period. Prediction times for \alg within parenthesis indicate those obtained on a CPU whereas those outside parentheses are times on a GPU.}
    \label{tab:baelines_eval}
      \centering
      \resizebox{\linewidth}{!}{
        \begin{tabular}{@{}l|cc|cc|c@{}}
        \toprule
        \textbf{Method} & \textbf{PSP@1} & \textbf{PSP@5} &  \textbf{P@1} & \textbf{P@5} &  \multicolumn{1}{c}{\begin{tabular}[c]{@{}c@{}}\textbf{Prediction}\\ \textbf{Time (ms)}\end{tabular}} \\
        \midrule
      \multicolumn{6}{c}{LF-AmazonTitles-131K}\\ \midrule					
\alg	 & \textbf{30.85}	 & \textbf{41.42}	 & \textbf{38.4}	 & \textbf{18.65}	 & \textbf{0.1} (1.15)\\
Astec	 & 29.22	 & 39.49	 & 37.12	 & 18.24	 & 2.34\\
AttentionXML	 & 23.97	 & 32.57	 & 32.25	 & 15.61	 & 5.19\\
MACH	 & 24.97	 & 34.72	 & 33.49	 & 16.45	 & 0.23\\
X-Transformer	 & 21.72	 & 27.09	 & 29.95	 & 13.07	 & 15.38\\
Siamese	 & 13.3	 & 13.36	 & 13.81	 & 5.81	 & 0.2\\
Parabel	 & 23.27	 & 32.14	 & 32.6	 & 15.61	 & 0.69\\
Bonsai	 & 24.75	 & 34.86	 & 34.11	 & 16.63	 & 7.49\\
DiSMEC	 & 25.86	 & 36.97	 & 35.14	 & 17.24	 & 5.53\\
PfastreXML	 & 26.81	 & 34.24	 & 32.56	 & 16.05	 & 2.32\\
XT	 & 22.37	 & 31.64	 & 31.41	 & 15.48	 & 9.12\\
Slice	 & 23.08	 & 31.89	 & 30.43	 & 14.84	 & 1.58\\
AnneXML	 & 19.23	 & 32.26	 & 30.05	 & 16.02	 & 0.11\\
\midrule
\multicolumn{6}{c}{LF-WikiSeeAlsoTitles-320K}\\ \midrule					
\alg	 & \textbf{16.73}	 & \textbf{21.01}	 & \textbf{25.14}	 & \textbf{12.86}	 & \textbf{0.09} (0.97)\\
Astec	 & 13.69	 & 17.5	 & 22.72	 & 11.43	 & 2.67\\
AttentionXML	 & 9.45	 & 11.73	 & 17.56	 & 8.52	 & 7.08\\
MACH	 & 9.68	 & 12.53	 & 18.06	 & 8.99	 & 0.52\\
X-Transformer	 & -	 & -	 & -	 & -	 & -\\
Siamese	 & 10.1	 & 9.59	 & 10.69	 & 4.51	 & 0.17\\
Parabel	 & 9.24	 & 11.8	 & 17.68	 & 8.59	 & 0.8\\
Bonsai	 & 10.69	 & 13.79	 & 19.31	 & 9.55	 & 14.82\\
DiSMEC	 & 10.56	 & 14.82	 & 19.12	 & 9.87	 & 11.02\\
PfastreXML	 & 12.15	 & 13.26	 & 17.1	 & 8.35	 & 2.59\\
XT	 & 8.99	 & 11.82	 & 17.04	 & 8.6	 & 12.86\\
Slice	 & 11.24	 & 15.2	 & 18.55	 & 9.68	 & 1.85\\
AnneXML	 & 7.24	 & 11.75	 & 16.3	 & 8.84	 & 0.13\\
\midrule
\multicolumn{6}{c}{LF-AmazonTitles-1.3M}\\ \midrule					
\alg	 & 22.07	 & 29.3	 & \textbf{50.67}	 & \textbf{40.35}	 & 0.16 (1.73)\\
Astec	 & 21.47	 & 27.86	 & 48.82	 & 38.44	 & 2.61\\
AttentionXML	 & 15.97	 & 22.54	 & 45.04	 & 36.25	 & 29.53\\
MACH	 & 9.32	 & 13.26	 & 35.68	 & 28.35	 & 2.09\\
X-Transformer	 & -	 & -	 & -	 & -	 & -\\
Siamese	 & -	 & -	 & -	 & -	 & -\\
Parabel	 & 16.94	 & 24.13	 & 46.79	 & 37.65	 & 0.89\\
Bonsai	 & 18.48	 & 25.95	 & 47.87	 & 38.34	 & 39.03\\
DiSMEC	 & -	 & -	 & -	 & -	 & -\\
PfastreXML	 & \textbf{28.71}	 & \textbf{32.51}	 & 37.08	 & 31.43	 & 23.64\\
XT	 & 13.67	 & 19.06	 & 40.6	 & 32.01	 & 5.94\\
Slice	 & 13.8	 & 18.89	 & 34.8	 & 27.71	 & 1.45\\
AnneXML	 & 15.42	 & 21.91	 & 47.79	 & 36.91	 & \textbf{0.12}\\
        \bottomrule
    \end{tabular}}
\end{table}

\textbf{Datasets}: Experiments were conducted on product-to-product and related-webpage recommendation datasets. These were short-text tasks with only the product/webpage titles being used to perform prediction. Of these, LF-AmazonTitles-131K, LF-AmazonTitles-1.3M, and LF-WikiSeeAlsoTitles-320K are publicly available at The Extreme Classification Repository~\cite{XMLRepo}. Results are also reported on two proprietary product-to-product recommendation datasets (LF-P2PTitles-300K and LF-P2PTitles-2M) mined from click logs of the Bing search engine, where a pair of products was considered similar if the Jaccard index of the set of queries which led to a click on them was found to be more than a certain threshold. We also considered some datasets' long text counterparts, namely LF-Amazon-131K and LF-WikiSeeAlso-320K, which contained the entire product/webpage descriptions. Note that LF-AmazonTitles-131K and LF-AmazonTitles-1.3M (as well as their long-text counterparts) are subsets of the standard AmazonTitles-670K and AmazonTitles-3M datasets respectively, and were created by restricting the label set to labels for which label-text was available. Please refer to Appendix~A.3 and Table~7 in the \suppl for dataset preparation details and dataset statistics.


\textbf{Baseline algorithms}: \alg was compared to leading deep extreme classifiers including the X-Transformer~\cite{Chang20}, Astec~\cite{Dahiya21}, XT~\cite{Wydmuch18}, AttentionXML~\cite{You18}, and MACH~\cite{Medini2019}, as well as standard extreme classifiers based on fixed or sparse BoW features including Bonsai~\cite{Khandagale19}, DiSMEC~\cite{Babbar17}, Parabel~\cite{Prabhu18b}, AnnexML~\cite{Tagami17}. Slice~\cite{Jain19}. Slice was trained with fixed FastText~\cite{Bojanowski17} features, while other methods used sparse BoW features. Unfortunately, GLaS~\citep{Guo2019} could not be included in the experiments as their code was not publicly available. Each baseline deep learning method was given a 12-core Intel Skylake 2.4 GHz machine with 4 Nvidia V100 GPUs. However, \alg was offered a 6-core Intel Skylake 2.4 GHz machine with a single Nvidia V100 GPU. A training timeout of 1 week was set for every method. Please refer to Table~9 in the \suppl for more details.

\clearpage

\begin{table}[ht]
    \caption{Results on proprietary product-to-product (P2P) recommendation datasets. C@20 denotes label coverage offered by the top 20 predictions of each method. \alg offers significantly better accuracies than all competing methods. Other methods such as AnnexML and DiSMEC did not scale with available resources within the timeout period.}
    \label{tab:p2p}
      \centering
      \resizebox{\linewidth}{!}{
        \begin{tabular}{@{}l|cc|cc|c@{}}
        \toprule
        \textbf{Method} & \textbf{PSP@1} & \textbf{PSP@5} &  \textbf{P@1} & \textbf{P@5} & \textbf{C@20} \\
        \midrule
        \multicolumn{6}{c}{LF-P2PTitles-300K}\\ \midrule					
        \alg	 & \textbf{42.43}	 & \textbf{62.3}	 & \textbf{47.17}	 & 22.69	 & 95.32\\
        Astec & 39.44 & 57.83 & 44.30 & 21.56 & 95.61\\
        Parabel	 & 37.26	 & 55.32	 & 43.14	 & 20.99	 & 95.59\\
        PfastreXML	 & 35.79	 & 49.9	 & 39.4	 & 18.77	 & 87.91\\
        Slice	 & 27.03	 & 34.95	 & 31.27	 & \textbf{25.19}	 & 95.06\\
        \midrule
        \multicolumn{6}{c}{LF-P2PTitles-2M}\\ \midrule					
        \alg	 & \textbf{36.65}	 & \textbf{45.15}	 & \textbf{40.27}	 & \textbf{31.45}	 & 93.08\\
        Astec	 & 32.75	 & 41	 & 36.34	 & 28.74	 & 95.3\\
        Parabel	 & 30.21	 & 38.46	 & 35.26	 & 28.06	 & 92.82\\
        PfastreXML	 & 28.84	 & 35.65	 & 30.52	 & 24.6	 & 88.05\\
        Slice	 & 27.03	 & 34.95	 & 31.27	 & 25.19	 & 93.43\\
        \bottomrule
    \end{tabular}}
\end{table}

\textbf{Evaluation}: Standard extreme classification metrics~\cite{Babbar19, Prabhu14, Prabhu18b, You18, Liu17}, namely Precision (P@$k$) and propensity scored precision (PSP@$k$) for $k= 1, 3, 5$ were used and are detailed in Appendix~A.4 in the \suppl.

\textbf{Hyperparameters}: \alg has two tuneable hyperparameters a) beam-width $B$ which determines the shortlist length $LB/K$ and b) token embedding dimension $D$. $B$ was chosen after concluding Module II training by setting a value that ensured a recall of $>85\%$ on the training set (note that choosing $B = K$ trivially ensures $100\%$ recall). Doing so did not require \alg to re-train Module II yet ensured a high quality shortlisting. Token embedding dimension $D$ was kept at 512 for larger datasets to improve the network capacity for large output spaces. For the small dataset LF-AmazonTitles-131K, clusters size $K$ was kept at $2^{15}$ and for other datasets it was kept at $2^{17}$. All other hyperparameters including learning rate, number of epochs were set to their default values across all datasets. Please refer to Table~8 in the \suppl for details.

\begin{figure}
    \centering
    \includegraphics[width=\linewidth]{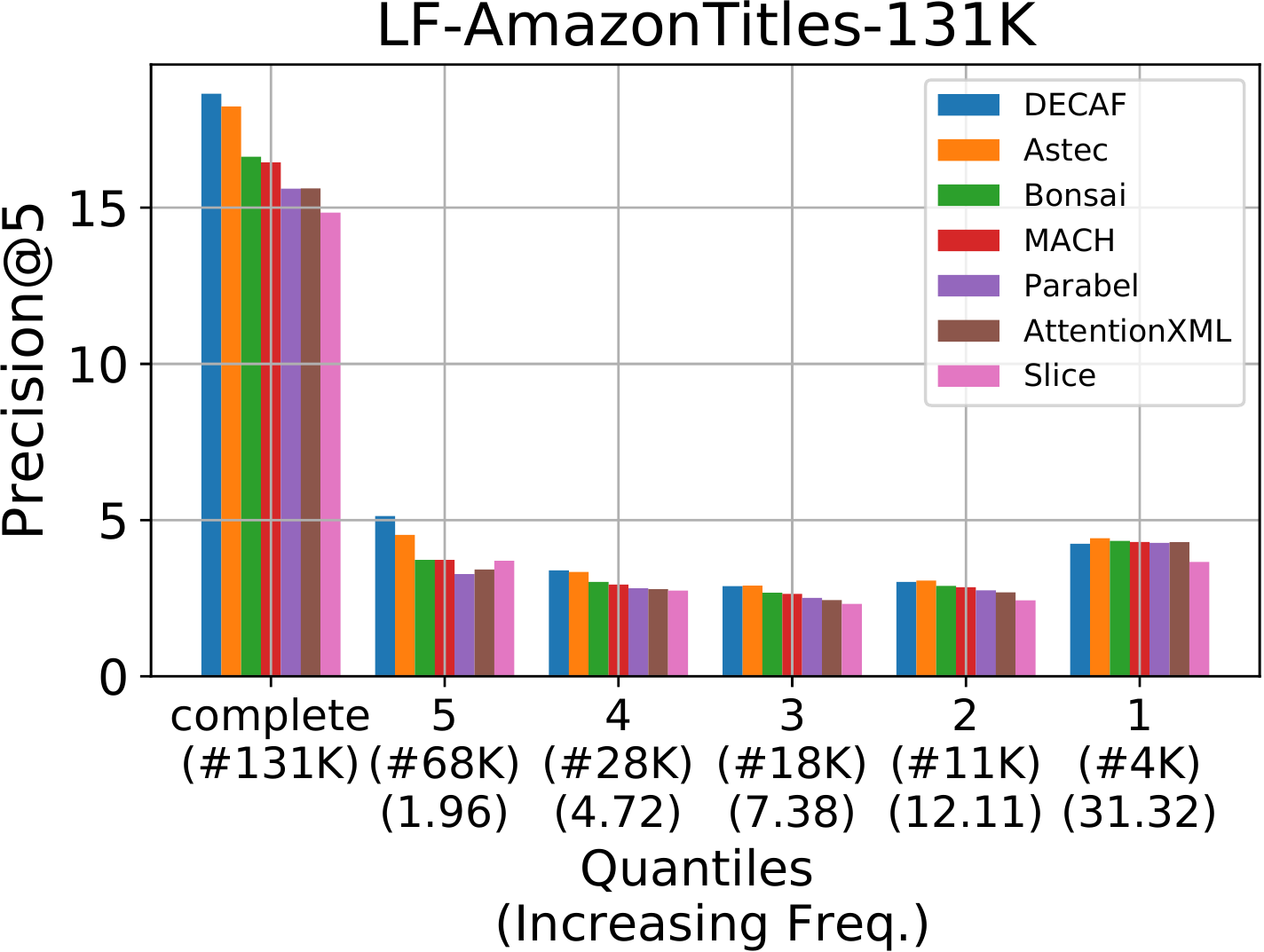}
    
    \vspace{0.5cm}
    
    \includegraphics[width=\linewidth]{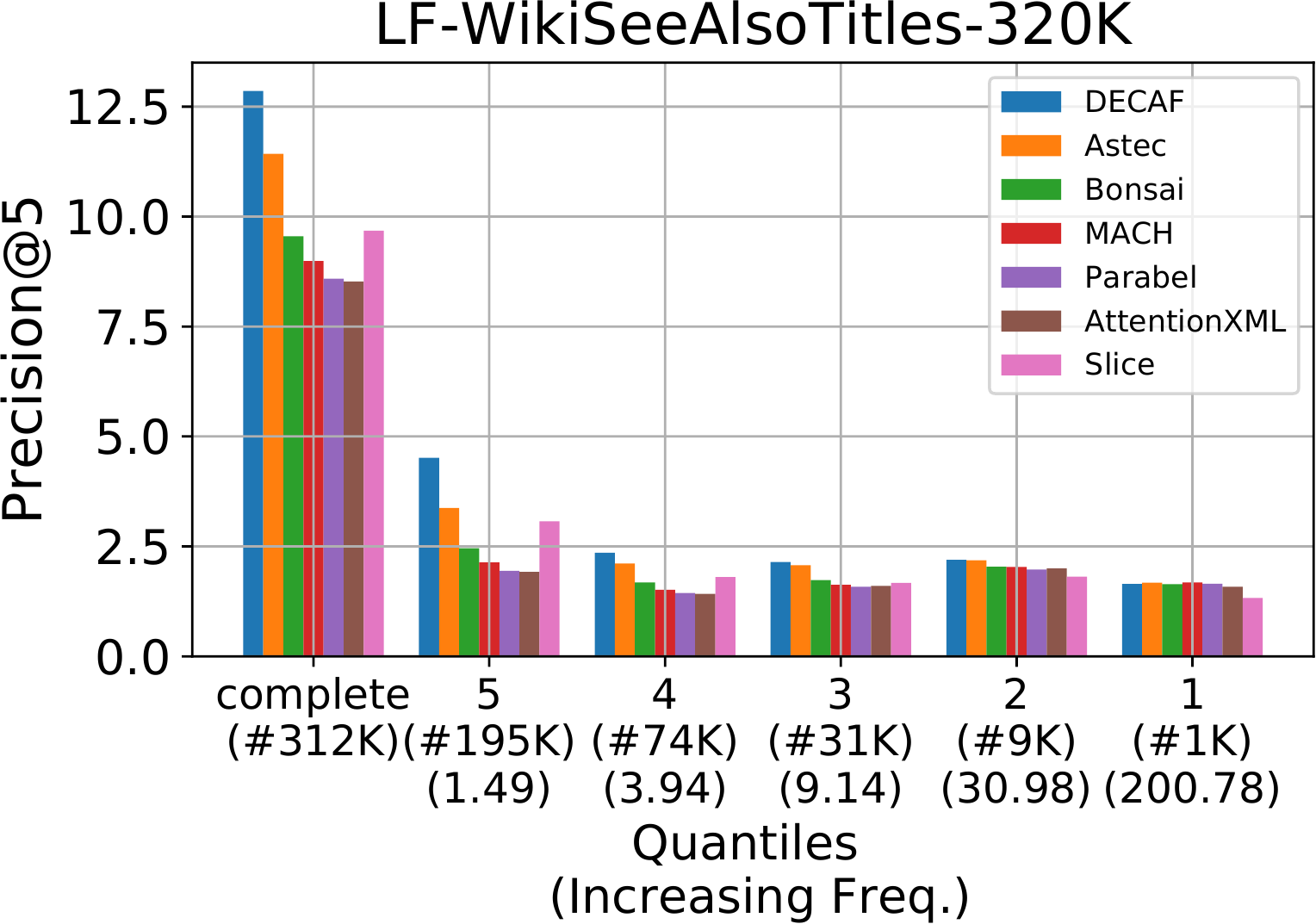}
    \caption{Quantile analysis of gains offered by \alg in terms of contribution to P@5. The label set was divided into 5 equi-voluminous bins with increasing label frequency. Quantiles increase in mean label frequency from left to right. \alg consistently outperforms other methods on all bins with the difference in accuracy being more prominent on bins containing data-scarce tail labels (e.g. bin 5).}
    \label{fig:sup:contrib}
\end{figure}

\textbf{Results on public datasets}: Table~\ref{tab:baelines_eval} compares \alg with leading XML algorithms on short-text product-to-product and related-webpage tasks. For details as well as results on long-text versions of these datasets, please refer to Table~9 in the \suppl. Furthermore, although \alg focuses on product-to-product applications, results on product-to-category style datasets such as product-to-category prediction on Amazon or article-to-category prediction on Wikipedia are reported in Table~10 in the \suppl. Parabel~\citep{Prabhu18}, Bonsai~\citep{Khandagale19}, AttentionXML~\citep{You18} and X-Transformer~\citep{Chang20} are the most relevant methods to \alg as they shortlist labels based on a tree learned in the label centroid space. \alg was found to be $4-10\%$ more accurate than methods such as Slice~\cite{Jain19}, PfastreXML~\cite{Jain17}, DiSMEC~\cite{Babbar17}, and AnnexML~\cite{Tagami17} that use fixed or pre-learnt features. This demonstrates that learning tasks-specific features can lead to significantly more accurate predictions. \alg was also compared with other leading deep learning based approaches like MACH~\citep{Medini2019}, and XT~\cite{Wydmuch18}. \alg could be up to $7\%$ more accurate while being more than 150$\times$ faster at prediction as compared to attention based models like X-Transformer and AttentionXML. \alg was also compared to Siamese networks that had similar access to label metadata as \alg. However, \alg could be up to $15\%$ more accurate than a Siamese network at an extreme scale. \alg was also compared to Astec~\citep{Dahiya21} that was specifically designed for short-text applications but does not utilize label metadata. \alg could be up to 3\% more accurate than Astec. This further supports \alg's claim of using label meta-data for improving prediction accuracy. Even on long-text tasks such as the LF-WikiSeeAlso-320K dataset (please refer to Table~9 in the \suppl), \alg can be more accurate in propensity scored metrics compared to the second best method AttentionXML, in addition to being vastly superior in terms of prediction time. This indicates the suitability of \alg's frugal architecture to product-to-product scenarios. The frugal architecture also allows \alg to make predictions on a CPU within a few milliseconds even for large datasets such as LF-AmazonTitles-1.3M while other deep extreme classifiers can take an order of magnitude longer time even on a GPU. \alg's prediction times on a CPU are reported within parentheses in Table~\ref{tab:baelines_eval}.

\textbf{Results on proprietary datasets}: Table~\ref{tab:p2p} presents results on proprietary product-to-product recommendation tasks (with details presented in Table~11 in the \suppl). \alg could easily scale to the LF-P2PTitles-2M dataset and be upto 2\% more accurate than leading XML algorithms including Bonsai, Slice and Parabel. Unfortunately, leading deep learning algorithms such as X-Transformer could not scale to this dataset within the timeout. \alg offers label coverage similar to state-of-the-art XML methods yet offers the best accuracy in terms of P@1. Thus, \alg's superior predictions do not come at a cost of coverage.

\begin{table}[t]
    \centering
    \caption{\alg's predictions on selected test points. Document and label names ending in ``$\dots$'' were abbreviated due to lack of space. Please refer to Table~12 in the \suppl for the complete table. Predictions in black and a non-bold/non-italic font were a part of the ground truth. Those in bold italics were part of the ground truth but never seen with other the ground truth labels in the training set i.e. had no common training points. Predictions in light gray were not a part of the ground truth. \alg's exploits label metadata to discover semantically correlated labels.}
    \label{tab:decaf_examples}
    \resizebox{\linewidth}{!}{
    \begin{tabular}{p{0.2\linewidth}|p{0.8\linewidth}}
        \toprule
        \textbf{Document} & \textbf{Top 5 predictions by \alg} \\
        \midrule
        {Panzer Dragoon Zwei} &  Panzer Dragoon, Action Replay Plus, Sega Saturn System - Video Game Console, \wpred{The Legend of Dragoon}, \emph{\textbf{Panzer Dragoon Orta}} \\
        \midrule
        {Wagner - Die Walkure $\dots$} & Wagner - Siegfried $\dots$, Wagner - Gotterdammerung $\dots$, Wagner - Der Fliegende Holländer (1986), \emph{\textbf{Wagner - Gotterdammerung $\dots$}}, \wpred{Seligpreisung} \\
        \midrule
        {New Zealand dollar} & \emph{\textbf{Economy of New Zealand}}, Cook Islands dollar, \wpred{Politics of New Zealand}, Pitcairn Islands dollar, \emph{\textbf{Australian dollar}} \\
        \bottomrule
    \end{tabular}
    }
\end{table}

\textbf{Analysis}: Table~\ref{tab:decaf_examples} shows specific examples of \alg predictions. \alg encourages collaborative learning among labels which allows it to predict the labels ``Australian dollar" and ``Economy of New Zealand'' for the document ``New Zealand dollar'' when other methods failed to do so. This example was taken from the LF-WikiseeAlsoTitles-320K dataset (please refer to Table~12 in the \suppl for details). It is notable that these labels do not share any common training instances with other ground truth labels but are semantically related nevertheless. \alg similarly predicted a rare label ``Panzer Dragoon Orta'' for the (video game) product ``Panzer Dragoon Zwei' whereas other algorithms failed to do so. To better understand the nature of \alg's gains, the label set was divided into five uniform bins (quantiles) based on frequency of occurrence in the training set. \alg's collaborative approach using label text in classifier learning led to gains in every quantile, the gains were more prominent on the data-scarce tail-labels, as demonstrated in Figure~\ref{fig:sup:contrib}.

\textbf{Incorporating metadata into baseline XML algorithms}: In principle, \alg's formulation could be deployed with existing XML algorithms wherever collaborative learning is feasible. Table \ref{tab:bowmeta} shows that introducing label text embeddings to the DiSMEC, Parabel, and Bonsai classifiers led to upto $1.5\%$ gain as compared to their vanilla counterparts that do not use label text. Details of these augmentations are given in Appendix~A.5 in the \suppl. Thus, label text inclusion can lead to gains for existing methods as well. However, \alg continues to be upto $7\%$ more accurate than even these augmented versions. This shows that \alg is more efficient at utilizing available label text.

\begin{table}
    \caption{Augmenting existing BoW-based XML methods by incorporating label metadata leads to $1.5\%$ increase in the accuracy as compared to base method. However, \alg could be up to $7\%$ more accurate compared to even these.}
    \label{tab:bowmeta}
      \centering
        \begin{tabular}{@{}l|cc|cc@{}}
        \toprule
         \textbf{Method} & \textbf{PSP@1}  & \textbf{PSP@5} & \textbf{P@1}  & \textbf{P@5} \\
        \midrule
        \multicolumn{5}{c}{LF-AmazonTitles-131K}\\ \midrule					
\alg	 & \textbf{30.85}	 & \textbf{41.42}	 & \textbf{38.4}	 & \textbf{18.65} \\
\midrule
Parabel	 & 23.27	 & 32.14	 & 32.6	 & 15.61 \\
Parabel + metadata	 & 25.89	 & 34.83	 & 33.6	 & 15.84 \\
\midrule
Bonsai	 & 24.75	 & 34.86	 & 34.11	 & 16.63 \\
Bonsai + metadata	 & 26.82	 & 36.63	 & 34.83	 & 16.67 \\
\midrule
DiSMEC	 & 26.25	 & 37.15	 & 35.14	 & 17.24 \\
DiSMEC + metadata	 & 27.19	 & 38.17	 & 35.52	 & 17.52 \\
    \midrule 
    \multicolumn{5}{c}{LF-WikiSeeAlsoTitles-320K}\\ \midrule					
\alg	 & \textbf{16.73}	 & \textbf{21.01}	 & \textbf{25.14}	 & \textbf{12.86} \\
\midrule
Parabel	 & 9.24	 & 11.8	 & 17.68	 & 8.59	 \\
Parabel + metadata	 & 12.96	 & 16.77	 & 20.69	 & 10.24 \\
\midrule
Bonsai	 & 10.69	 & 13.79	 & 19.31	 & 9.55	 \\
Bonsai + metadata	 & 13.63	 & 17.54	 & 21.61	 & 10.72 \\
\midrule
DiSMEC	 & 10.56	 & 14.82	 & 19.12	 & 9.87	\\
DiSMEC + metadata	 & 12.46	 & 15.9	 & 20.74	 & 10.29\\
\bottomrule
    \end{tabular}
\end{table}

\begin{table}
\caption{Using strategies used by existing XML algorithms for shortlisting labels instead of $\cS$ hurts both both shortlist recall (R@20) and final prediction accuracy (P@k, PSP@k).}
    \label{tab:sub:xmlclass}
    \centering
    \begin{tabular}{@{}l|cc|cc|c@{}}
        \toprule
        \textbf{Method}
        & \textbf{PSP@1}  & \textbf{PSP@5} & \textbf{P@1}  & \textbf{P@5} &\textbf{R@20}\\
        \midrule
\multicolumn{6}{c}{LF-AmazonTitles-131K}\\ \midrule					
\alg	 & \textbf{30.85}	 & \textbf{41.42}	 & \textbf{38.4}	 & \textbf{18.65}	 & \textbf{55.86} \\
\midrule
+ HNSW Shortlist	 & 29.55	 & 39.17	 & 36.7	 & 17.78	 & 48.82 \\
+ Parabel Shortlist	 & 24.88	 & 31.21	 & 32.13	 & 14.73	 & 39.36 \\
\midrule
\multicolumn{6}{c}{LF-WikiSeeAlsoTitles-320K}\\ \midrule			
\alg	 & \textbf{16.73}	 & \textbf{21.01}	 & \textbf{25.14}	 & \textbf{12.86}	 & \textbf{37.53} \\
\midrule
+ HNSW Shortlist	 & 15.68	 & 19.38	 & 23.84	 & 12.11	 & 30.26 \\
+ Parabel Shortlist	 & 13.17	 & 15.09	 & 21.18	 & 10.05	 & 23.91 \\
        \bottomrule
    \end{tabular}
\end{table}

\begin{table}
    \caption{Analyzing the impact for alternative design and algorithmic choices for DECAF’s components.}
    \label{tab:combouv}
      \centering
      \resizebox{\linewidth}{!}{
        \begin{tabular}{@{}l|cc|cc|c@{}}
        \toprule
        Component & \textbf{PSP@1}  & \textbf{PSP@5} & \textbf{P@1}  & \textbf{P@5} & \textbf{R@20} \\
        \midrule
\multicolumn{6}{c}{LF-AmazonTitles-131K}\\ \midrule					
\alg	 & \textbf{30.85}	 & \textbf{41.42}	 & \textbf{38.4}	 & \textbf{18.65}	 & \textbf{55.86} \\
\alg-FFT	 & 25.5	 & 33.38	 & 32.42	 & 15.43	 & 47.23 \\
\alg-8K	 & 29.07	 & 38.7	 & 36.29	 & 17.52	 & 51.65 \\
\alg-no-init	 & 29.86	 & 41.04	 & 37.79	 & 18.57	 & 55.75 \\
\alg-$\hat\vz^1$	 & 28.02	 & 38.38	 & 33.5	 & 17.09	 & 53.83 \\
\alg-$\hat\vz^2$	 & 27.32	 & 38.05	 & 36	 & 17.65	 & 52.2 \\
\algl	 & 29.75	 & 40.36	 & 37.26	 & 18.29	 & 55.25 \\
        \midrule
        \multicolumn{6}{c}{LF-WikiSeeAlsoTitles-320K}\\ \midrule					
\alg	 & 16.73	 & 21.01	 & \textbf{25.14}	 & \textbf{12.86}	 & \textbf{37.53} \\
\alg-FFT	 & 13.91	 & 17.3	 & 21.72	 & 11	 & 32.58 \\
\alg-8K	 & 14.55	 & 17.38	 & 22.41	 & 10.96	 & 30.21 \\
\alg-no-init	 & 15.09	 & 19.47	 & 23.81	 & 12.25	 & 36.18 \\
\alg-$\hat\vz^1$	 & \textbf{18.04}	 & \textbf{21.48}	 & 24.54	 & 12.55	 & 37.33 \\
\alg-$\hat\vz^2$	 & 11.55	 & 15.24	 & 20.82	 & 10.53	 & 29.72 \\
\algl	 & 16.59	 & 20.84	 & 24.87	 & 12.78	 & 37.24 \\
\bottomrule
        \end{tabular}}
\end{table}

\begin{figure}
    \centering
    \begin{minipage}{\linewidth}
		\centering
	   \includegraphics[width=0.65\linewidth]{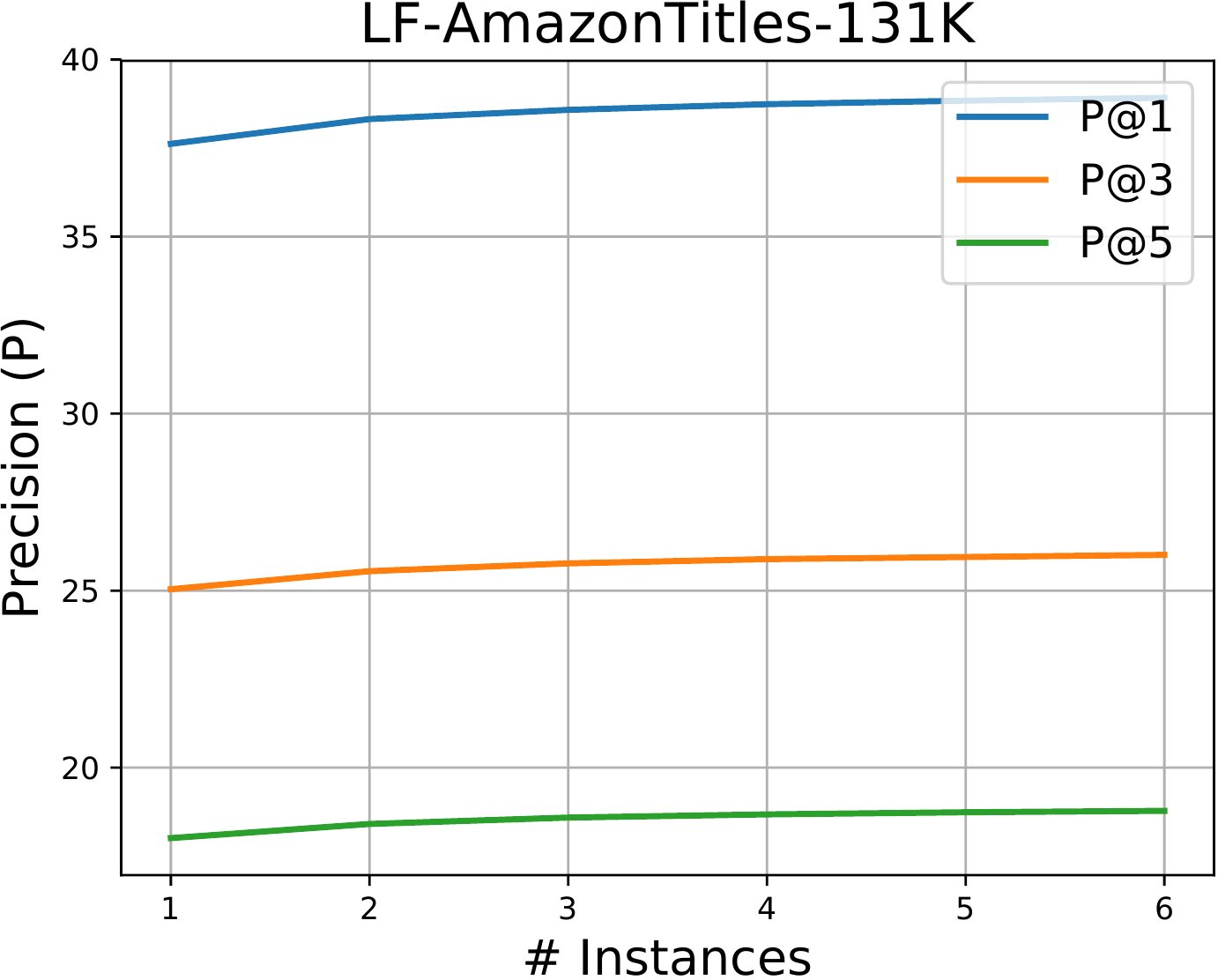}
    \caption{Impact of the number of instances in \alg's ensemble on performance on the LF-AmazonTitles-131K dataset. \alg offers maximum benefits using a small ensemble of 3 instances after which benefits taper off.}
    \label{fig:numlearner}
	\end{minipage}
	\begin{minipage}{\linewidth}
		\centering
	   \includegraphics[width=0.65\linewidth]{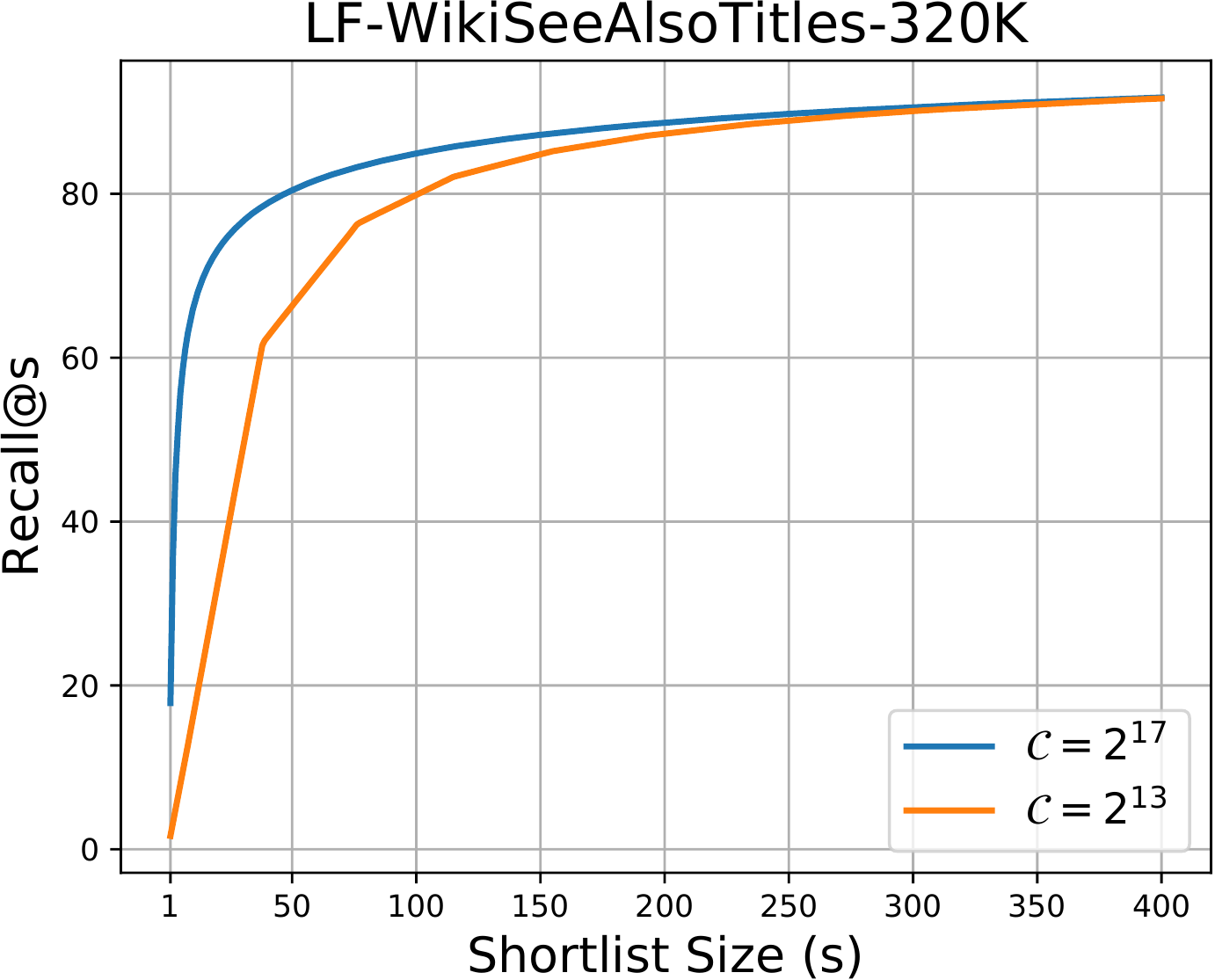}
    \caption{A comparison of recall when using moderate or large fanout on the LF-WikiSeeAlso-320K dataset. The x-axis represents various values of beam-width $B$ and training recall offered by each. A large fanout offers superior recall with small beam width, and hence small shortlists lengths.}
    \label{fig:recall_clusters}
	\end{minipage}
\end{figure}

\textbf{Shortlister}: \alg's shortlister distinguishes itself from previous shortlisting strategies \cite{Chang20,Khandagale19,You18,Prabhu18b} in two critical ways. Firstly, \alg uses a massive fanout of $K = 2^{17} \approx 130$K clusters whereas existing approaches either use much fewer (upto 8K) clusters \cite{Chang20,Bhatia15} or use hierarchical clustering with a small fanout (upto 100) at each node \cite{Khandagale19,You18}. Secondly, in contrast to other methods that create shortlists from generic embeddings (e.g. bag-of-words or FastText~\citep{Joulin17}), \alg fine-tunes its shortlister in Module II using task-specific embeddings learnt in Module I. Tables \ref{tab:sub:xmlclass} and \ref{tab:combouv} show that \alg's shortlister offers much better performance than shortlists computed using a small fanout or else computed using ANNS-based negative sampling \cite{Jain19}. Fig~\ref{fig:recall_clusters} shows that a large fanout offers much better recall even with small shortlist lengths than if using even moderate fanouts e.g. $K = 8$K.

\textbf{Ablation}: As described in Section \ref{sec:method}, the training pipeline for \alg is divided into 4 modules mirroring the DeepXML pipeline \cite{Dahiya21}. Table~\ref{tab:combouv} presents the results of extensive experiments conducted to analyze the optimality of algorithmic and design choices made in these modules. We refer to Appendix~A.5 in the \suppl for details. \textbf{a)} To assess the utility of learning task-specific token embeddings in Module I, a variant \alg-FFT was devised that replaced these with pre-trained FastText embeddings: \alg outperforms \alg-FFT by 6\% in PSP@1 and 3.5\% in P@1. \textbf{b)} To assess the impact of a large fanout while learning the shortlister, a variant \alg-8K was trained with a smaller fanout of $K = 2^{13} \approx 8$K clusters that is used by methods such as AttentionXML and X-Transformer. Restricting fanout was found to hurt accuracy by 3\%. This can be attributed to the fact that the classifier's final accuracy depends on the recall of the shortlister (see Theorem~\ref{thm:thm}). Fig.~\ref{fig:recall_clusters} indicates that using $K = 2^{13}$ results in significantly larger shortlist lengths (upto $2\times$ larger) being required to achieve the same recall as compared to using $K = 2^{17}$. Large shortlists make Module IV training and prediction more challenging, especially for large datasets involving millions of labels, thereby making a large fan-out $K$ more beneficial. \textbf{c)} Approaches other than \alg's shortlister $\cS$ were considered for shortlisting labels, such as nearest neighbor search using HNSW \cite{Jain19} or PLTs with small fanout such as Parabel \cite{Prabhu18b} learnt over dense document embeddings. Table \ref{tab:sub:xmlclass} shows that both alternatives lead to significant loss, upto 15\% in recall, as compared to that offered by $\cS$. These sub-optimal shortlists eventually hurt final prediction which could be 2\% less accurate as compared to \alg. \textbf{d)} To assess the importance of label classifier initialization in Module III, a variant \alg-no-init was tested which initialized $\hat\vz^2_l$ randomly instead of with $\vE\vz_l$. \alg-no-init was found to offer 1-1.5\% less PSP@1 than \alg, therefore indicating importance of proper initialization in Module III. \textbf{e)} Modules II and IV learn OvA classifiers as a combination of the label embedding vector and a refinement vector. To investigate the need for both components, Table~\ref{tab:combouv} considers two \alg variants: the first variant, named \alg-$\hat\vz^1$, discards the refinement vector in both modules i.e. using $\vw_l = \hat\vz^1_l$ and $\vh_m = \hat\vu^1_m$ whereas the second variant, named \alg-$\hat\vz^2$, rejects the label embedding component altogether and learns the OvA classifers from scratch using only the refinement vector i.e. using $\vw_l = \hat\vz^2_l$ and $\vh_m = \hat\vu^2_m$. Both variants take a hit of up to 5\% in prediction accuracy as compared to \alg. Incorporating label-text in the classifier is critical to achieve superior accuracies. \textbf{f)} Finally, to assess the utility of fine-tuning token embeddings in each successive module, a frugal version \algl was considered which freezes token embeddings after Module I and shares token embeddings among the three instances in its ensemble. \algl offers 0.5-1\% loss in performance as compared to \alg but is noticeably faster at training.
\section{Conclusion}
\label{sec:conc}
This paper demonstrated the impact of incorporating label metadata in the form of label text in offering significant performance gains on several product-to-product recommendation tasks. It proposed the \alg algorithm that uses a frugal architecture, as well as a scalable prediction pipeline, to offer predictions that are up to 2-6\% more accurate, as well as an order of magnitude faster, as compared to leading deep learning-based XML algorithms. \alg offers millisecond-level prediction times on a CPU making it suitable for real-time applications such as product-to-product recommendation tasks. Future directions of work include incorporating other forms of label metadata such as label-correlation graphs, as well as diverse embedding architectures.

\begin{acks}
The authors thank the IIT Delhi HPC facility for computational resources. AM is supported by a Google PhD Fellowship.
\end{acks}

\bibliographystyle{ACM-Reference-Format}
\bibliography{ms}

\clearpage
\appendix

\onecolumn

\section{Appendix}
In this supplementary material, we present various details omitted from the main text due to lack of space, including a proof of Thm~\ref{thm:thm}, a detailed analysis of the time complexity of the various modules in the training and prediction pipelines of \alg, details of the datasets and evaluation metrics used in the experiments, further clarifications about how some ablation experiments were carried out, as well as additional experimental results including a subjective comparison of the prediction quality of \alg and various competitors on handpicked recommendation examples.

\subsection{Proof of Theorem~\ref{thm:thm}}
\label{app:proof}
We recall from the main text that $\cL(\Theta)$ denotes the original likelihood expression and $\tilde\cL(\Theta \cond \cS)$ denotes the approximate likelihood expression that incorporates the shortlister $\cS$. Both expression are reproduced below for sake of clarity.
\begin{align*}
\cL(\Theta) &= \frac1{NL}\sum_{i \in [N]}\sum_{l \in [L]}\ell_{il}(\Theta)\\
\tilde\cL(\Theta \cond \cS) &= \frac{K}{NLB}\sum_{i \in [N]}\sum_{l \in \cS(\hat\vx_i)}\ell_{il}(\Theta)
\end{align*}

\begin{theorem}[Theorem~\ref{thm:thm} Restated]
\label{thm:thm-restated}
Suppose the training data has label sparsity at rate $s$ i.e. $\sum_{i \in [N]}\sum_{l \in [L]} \bI\bc{y_{il} = +1} = s\cdot NL$ and the shortlister offers a recall rate of $r$ on the training set i.e. $\sum_{i \in [N]}\sum_{l \in \cS(\hat\vx_i)}\bI\bc{y_{il} = +1} = rs\cdot NL$. Then if $\hat\Theta$ is obtained by optimizing the approximate likelihood function $\tilde\cL(\Theta \cond \cS)$, then the following always holds
\[
\cL(\hat\Theta) \leq \min_{\Theta} \cL(\Theta) + \bigO{s(1-r)\ln\br{\frac1{s(1-r)}}}.
\]
\end{theorem}

Below we prove the above claimed result. For the sake of simplicity, let $\Theta^\ast = \argmin_\Theta\ \cL(\Theta)$ denote the optimal model that could have been learnt using the original likelihood expression. As discussed in Sec~\ref{sec:method}, OvA methods with linear classifiers assume a likelihood decomposition of the form $\P{\vy_i \cond \vx_i, \Theta} = \prod_{l=1}^L\P{y_{il} \cond \hat\vx_i, \vw_l} = \prod_{l=1}^L\br{1 + \exp\br{y_{il}\cdot\ip{\hat\vx_i}{\vw_l}}}^{-1}$ where $\hat\vx_i = \text{ReLU}(\cE(\vx_i))$ is the document embedding obtained using token embeddings $\vE$ and embedding block parameters taken from $\Theta$, and $\vw_l$ is the label classifier obtained as shown in Fig~\ref{fig:embedding}. Thus, for a label-document pair $(i,l) \in [N] \times [L]$, the model posits a likelihood
\[
\P{y_{il} \cond \hat\vx_i, \vw_l} = \br{1 + \exp\br{y_{il}\cdot\ip{\hat\vx_i}{\vw_l}}}^{-1}
\]
However, in the presence of a shortlister $\cS$, the above model fails to hold since for a document $i$, a label $l \notin \cS(\hat\vx_i)$ is never predicted. This can cause a catastrophic collapse of the model likelihood if even a single positive label fails to be shortlisted by the shortlister, i.e. if the shortlister admits even a single false negative. To address this, and allow \alg to continue working with shortlisters with high but still imperfect recall, we augment the likelihood model as follows
\[
\P{y_{il} \cond \hat\vx_i, \vw_l} = \begin{cases}
\br{1 + \exp\br{y_{il}\cdot\ip{\hat\vx_i}{\vw_l}}}^{-1} & l \in \cS(\hat\vx_i)\\
y_{il}\br{\eta - \frac12} + \frac12 & l \notin \cS(\hat\vx_i)
\end{cases},
\]
where $\eta \in (0,1]$ is some default likelihood value assigned to positive labels that escape shortlisting (recall that $y_{il} \in \bc{-1,+1}$). Essentially, for non-shortlisted labels, we posit their probability of being relevant as $\eta$. The value of $\eta$ will be tuned later.

Note that we must set $\eta \ll 1$ so as to ensure that these default likelihood scores do not interfere with the prediction pipeline which discards non-shortlisted labels. We will see that our calculations do result in an extremely small value of $\eta$ as the optimal value. However, also note that we cannot simply set $\eta = 0$ since that would lead to a catastrophic collapse of the model likelihood to zero if the shortlister has even one false negative. Although our shortlister does offer good recall even with shortists of small length (e.g. 85\% with a shortlist of length $\approx 200$), demanding 100\% recall would require exorbitantly large beam sizes that would slow down prediction greatly. Thus, it is imperative that the augmented likelihood model itself account for shortlister failures.

To incorporate the above augmentation, we also redefine our log-likelihood score function to handle document-label pairs $(i,l) \in [N] \times [L]$ such that $l \notin \cS(\hat\vx_i)$
\[
\ell_{il}(\Theta \cond \cS) = \begin{cases}
\ln\br{1 + \exp\br{y_{il}\cdot\ip{\hat\vx_i}{\vw_l}}} & l \in \cS(\hat\vx_i)\\
-\ln\br{y_{il}\br{\eta - \frac12} + \frac12} & l \notin \cS(\hat\vx_i)
\end{cases},
\]
Note the negative sign in the second case since $\ell_{ij}$ is the negative log-likelihood expression. We will also benefit from defining the following \emph{residual} loss term
\[
\Delta(\Theta \cond \cS) = \sum_{i \in [N]}\sum_{l \notin \cS(\hat\vx_i)}\ell_{il}(\Theta)
\]
Note that $\Delta$ simply sums up loss terms corresponding to all labels omitted by the shortlister. We will establish the result claimed in the theorem by comparing the performance offered by $\hat\Theta$ and $\Theta^\ast$ on the loss terms given by $\tilde\cL$ and $\Delta$. Note that for any $\Theta$ we always have the following decomposition
\[
\cL(\Theta) = \frac1{NL}\br{\frac{NLB}K\cdot\tilde\cL(\Theta \cond \cS) + \Delta(\Theta \cond \cS}
\]
\newcommand{\fnr}{\text{FPR}}
\newcommand{\tnr}{\text{TNR}}
Now, since $\hat\Theta$ optimizes $\tilde\cL$, we have $\tilde\cL(\hat\Theta \cond \cS) \leq \tilde\cL(\Theta^\ast \cond \cS)$ which settles the first term in the above decomposition. To settle the second term, we note that as per the recall $r$ and label sparsity $s$ terms defined in the statement of the theorem, the number of positive labels not shortlisted by the shortlister $\cS$ throughout the dataset is $\fnr\cdot NL$ where $\fnr = (1-r)s$ is the false negative rate of the shortlister. Similarly, the number of negative labels not shortlisted by the shortlister throughout the dataset by $(L-B)N$ can be seen to be $\tnr\cdot NL$ where $\tnr = \br{(1-s) - \frac BK + rs}$ is the true negative rate of the shortlister. This gives us
\[
\Delta(\hat\Theta \cond \cS) = \br{\fnr\cdot\ln\frac1\eta + \tnr\cdot\ln\frac1{1-\eta}}\cdot NL
\]
It is easy to see that the optimal value of $\eta$ for the above expression is $\eta = \frac{\fnr}{\fnr + \tnr}$. For example, in the LF-WikiSeeAlsoTitles-320K dataset, which has $s \approx 6.75 \times 10^{-6}, r \approx 0.85, B = 160, K = 2^{17}$, this gives a value of $\fnr \approx 1.01 \times 10^{-6}, \tnr \approx 0.999$ which gives $\eta \approx 1.01 \times 10^{-6}$. This confirms that the augmentation indeed does not interfere with the prediction pipeline and labels not shortlisted can be safely ignored. However, moving on and plugging this optimal value of $\eta$ into the expression tells us that
\[
\Delta(\hat\Theta \cond \cS) = \br{\frac\fnr\tnr\ln\br{1 + \frac\tnr\fnr} + \ln\br{1+\frac\fnr\tnr}}\cdot NL.
\]
Since $\tnr \rightarrow 1$ (for example, we saw $\tnr \approx 0.999$ above), we simplify this to $\frac\fnr\tnr = \bigO{\fnr}$ and use the inequality $\ln(1+v) \leq v$ for all $v > 0$ to conclude that $\Delta(\hat\Theta \cond \cS) \leq \bigO{\fnr\ln\frac1\fnr + \fnr} = \bigO{s(1-r)\ln\br{\frac1{s(1-r)}}}$. Using $\Delta(\Theta^\ast \cond \cS) \geq 0$ settles the second term in the decomposition by establishing that $\Delta(\hat\Theta \cond \cS) - \Delta(\Theta^\ast \cond \cS) \leq \bigO{s(1-r)\ln\br{\frac1{s(1-r)}}}\cdot NL$. Combining the two terms in the decomposition above gives us
\[
\cL(\hat\Theta) - \cL(\Theta^\ast) = \frac1{NL}\br{\frac{NLB}K\cdot(\tilde\cL(\Theta \cond \cS) - \tilde\cL(\Theta^\ast \cond \cS)) + (\Delta(\Theta \cond \cS) - \Delta(\Theta^\ast \cond \cS))} \leq \bigO{s(1-r)\ln\br{\frac1{s(1-r)}}},
\]
which finishes the proof of the theorem.

We conclude this discussion by noting that since $\cL$ and $\tilde\cL$ are non-convex objectives due to the non-linear architecture encoded by the model parameters $\Theta$, it may not be able to solve these objectives optimally in practice. Thus, in practice, all we may be ensure is that
\[
\tilde\cL(\hat\Theta \cond \cS) \leq \min_\Theta \tilde\cL(\Theta \cond \cS) + \epsilon_{\text{opt}}
\]
where $\epsilon_{\text{opt}}$ is the sub-optimality in optimizing the objective $\tilde\cL$ due to factors such as sub-optimal initialization, training, premature termination, etc. It is easy to see that the main result of the theorem continues to hold since we now have $\tilde\cL(\hat\Theta \cond \cS) \leq \tilde\cL(\Theta^\ast \cond \cS) + \epsilon_{\text{opt}}$ which gives us the modified result as follows
\[
\cL(\hat\Theta) \leq \min_\Theta \cL(\Theta) + \bigO{s(1-r)\ln\br{\frac1{s(1-r)}}} + \frac BK\epsilon_{\text{opt}}.
\]

\newcommand{\vdoc}{\hat V_x}
\newcommand{\vlab}{\hat V_y}
\newcommand{\hL}{\hat L}
\newcommand{\hN}{\hat N}

\allowdisplaybreaks

\subsection{Time Complexity Analysis for \alg}
\label{app:complexity}
In this section, we discuss the time complexity of the various modules in \alg, as well as derive the prediction and training complexities.

\textbf{Notation}: Recall from Section~\ref{sec:method} that \alg learns $D$-dimensional representations for all $V$ tokens ($\ve_t, t \in [V]$), that are used to create embeddings for all $L$ labels $\hat\vz^1_l, l \in [L]$, and all $N$ training documents $\hat\vx_i, i \in [N]$. We introduce some additional notation to facilitate the discussion: we use $\vdoc$ to denote the average number of unique tokens present in a document i.e. $\vdoc = \frac1N\sum_{i=1}^N\norm{\vx_i}_0$ where $\norm{\cdot}_0$ is the sparsity ``norm'' that gives the number of non-zero elements in a vector. We similarly use $\vlab = \frac1L\sum_{l=1}^L\norm{\vz_l}_0$ to denote the average number of tokens in a label text. Let $\hL =  \frac1N\sum_{i=1}^N\norm{\vy_i}_0$ denote the average number of labels per document and also let $\hN = \frac{N\hL}L$ denote the average number of documents per label. We also let $M$ denote the mini-batch size (\alg used $M = 255$ for all datasets -- see Table~\ref{tab:hyperparameters}).

\textbf{Embedding Block}: Given a text with $\hat V$ tokens, the embedding block requires $\hat VD$ operations to aggregate token embeddings and $D^2 + 3D$ operations to execute the residual block and the combination block, for a total of $\bigO{\hat VD + D^2}$ operations. Thus, to encode a label (respectively document) text, it takes $\bigO{\vlab D + D^2}$ (respectively $\bigO{\vdoc D + D^2}$) operations on average.

\textbf{Prediction}: Given a test document, assuming that it contain $\vdoc$ tokens, embedding takes $\bigO{\vdoc D + D^2}$ operations, executing the shortlister by identifying the top $B$ clusters takes $\bigO{KD + K\log K}$ operations. These clusters contain a total of $\frac{LB}K$ labels. The ranker takes $\bigO{\frac{LB}KD + \frac{LB}K\log\br{\frac{LB}K}}$ operations to execute the $\frac{LB}K$ OvA linear models corresponding to these shortlisted labels to obtain the top-ranked predictions. Thus, prediction takes $\bigO{\vdoc D + D^2 + KD + K\log K} = \bigO{KD}$ time since usually $\frac{LB}K \leq K, \vdoc \leq K$ and $\log K \leq D \leq K$.

\textbf{Module I Training}: Creation of all $L$ label centroids $\vc_l$ takes $\bigO{L\hN\vdoc}$ time. These centroids are $\bigO{\hN\vdoc}$-sparse on average. Clustering these labels using hierarchical balanced binary clustering for $\log K$ levels to get $K$ balanced clusters takes time $\bigO{L\hN\vdoc\log K}$. Computing meta label text representations $\vu_m$ for all meta labels takes $\bigO{L\vlab}$ time. The vectors $\vu_m$ are $\frac{\vlab L}K$-sparse on average. To compute the complexity of learning the $K$ OvA meta-classifiers, we calculate below the cost of a single back-propagation step when using a mini-batch of size $M$. Computing the document and meta-label features of all $M$ documents in the mini-batch and $K$ meta-labels takes on average $\bigO{(D^2 + \vdoc D)M}$ and $\bigO{\br{D^2 + \frac{\vlab L}K\cdot D}K}$ time respectively. Computing the scores for all the OvA meta classifiers for all documents in the mini-batch takes $\bigO{MKD}$ time. Overestimating that the $K$ meta label texts together cover all $V$ tokens, updating the residual layer parameters $\vR$, the combination block parameters, and the token embeddings $\vE$ using back-propagation takes at most $\bigO{(D^2 + V)MK}$ time.



\textbf{Module II Training}: Recreating all $L$ label centroids $\vc_l$ now takes $\bigO{L\hN\vdoc D}$ time. Clustering the labels takes time $\bigO{LD\log K}$. Computing document features in a mini-batch of size $M$ takes $\bigO{(\vdoc D + D^2)M}$ time as before. Computing the meta-label representations $\hat\vu^1_m$ for all $K$ meta-labels now takes $\bigO{(\vlab D + D^2)L}$ time. Computing the scores for all the OvA meta classifiers for all documents in the mini-batch takes $\bigO{MKD}$ time as before. Next, updating the model parameters as well as the refinement vectors $\hat\vu^2_m, m \in [K]$ takes at most $\bigO{(D^2 + V)MK}$ time time as before. The added task of updating $\hat\vu^2_m$ does not affect the asymptotic complexity of this module. Generating the shortlists for all $N$ training points is essentially a prediction step and takes $\bigO{NKD}$ time.

\textbf{Module II Initializations}: Model parameter initializations take $\bigO{D^2}$ time. Initializing the refinement vectors $\hat\vz^2_l$ takes $\bigO{L\vlab D}$ time. 

\textbf{Module IV Training}: Given the shortlist of $LB/K$ labels per training point generated in Module II, training the OvA classifiers by fine-tuning the model parameters and learning the refinement vectors $\hat\vz^2_l, l \in [L]$ is made much less expensive than $\bigO{NLD}$. Computing document features in a mini-batch of size $M$ takes $\bigO{(\vdoc D + D^2)M}$ time as before. However, label representations $\hat\vz^1_l$ of only shortlisted labels need be computed. Since there are atmost $\br{\frac{LB}K + \hL}M$ of them (accounting for hard negatives and all positives), this takes $\bigO{(\vlab D + D^2)M\br{\frac{LB}K + \hL}}$ time. Next, updating the model parameters as well as the refinement vectors $\hat\vz^2_l$ for shortlisted takes at most $\bigO{(D^2 + (\vdoc +  \vlab)D)M\br{\frac{LB}K + \hL}}$ time. This can be simplified to $\bigO{M\br{\frac{LB}K + \hL}D^2} = \bigO{MD^2\log^2L}$ time per mini-batch since $\vdoc, \vlab \leq D$, usually $\hL \leq \bigO{\log L}$ and \alg chooses $\frac BK \leq \bigO{\frac{\log^2L}L}$ for large datasets such as LF-AmazonTitles-1.3M and LF-P2PTitles-2M (see Table~\ref{tab:hyperparameters}), thus ensuring an OvA training time that scales at most as $\log^2L$ with the number of labels.

\subsection{Dataset Preparation and Evaluation Details}
\label{app:dataprep}
Train-test splits were generated using a random 70:30 split keeping only those labels that have at least 1 test as well as 1 train point. For sake of validation, 5\% of training data points were randomly sampled.

\textbf{Reciprocal pair removal}:  It was observed that in certain datasets, documents were mapped to themselves. For instance, the product with title ``Dinosaur'' was tagged with the label ``Dinosaur'' itself in the LF-AmazonTitles-131K dataset. Algorithms could achieve disproportionately high P@$1$ by making such trivial predictions without learning anything useful. Additionally, in product-to-product and related webpage recommendation tasks, both documents and labels come from the same set/universe. This allows for \emph{reciprocal pairs} to exist where a data point has document A and label B in its ground truth but a separate data point has document B and label A in its ground truth. We affectionately call these AB and BA pairs respectively. If these pairs are split across train and test sets, an algorithm could simply memorize the AB pair while training and predict the BA pair during testing to achieve very high P@1. Moreover, such predictions did not add to the quality of predictions in real-life applications. Hence, methods were not rewarded for making such trivial predictions. Table~\ref{tab:baelines_eval} reports numbers as per this very evaluation strategy.
Additionally, coverage (C@20) is reported in Table~\ref{tab:p2p} to verify that prediction accuracy is not being achieved at the expense of label coverage.

\subsection{Evaluation metrics}
\label{sup:eval}
Performance was evaluated using precision@$k$ and nDCG@$k$ metrics. Performance was also evaluated using propensity scored metrics, namely propensity scored precision@$k$ and nDCG@$k$ (with $k$ = 1, 3 and 5) for extreme classification. The propensity scoring model and values available on The Extreme Classification Repository~\citep{XMLRepo} were used for the publicly available datasets. For the proprietary datasets, the method outlined in \cite{Jain16} was used. For a predicted score vector $\hat{\mathbf{y}} \in R^L$ and ground truth label vector $\mathbf{y} \in \{0, 1\}^L$, the metrics are defined below. In the following, $p_l$ is propensity score of the label $l$ as proposed in~\citep{Jain16}.
\begin{flalign}
P@k&= \frac{1}{k} {\sum_{l \in rank_k(\hat{\mathbf{y}})}} y_l &
PSP@k&= \frac{1}{k} {\sum_{l \in rank_k(\hat{\mathbf{y}})}} \frac{y_l}{p_l} & \nonumber\\
DCG@k&= \frac{1}{k} {\sum_{l \in rank_k(\hat{\mathbf{y}})}} \frac{y_l}{\log(l+1)} \nonumber
& PSDCG@k&= \frac{1}{k} {\sum_{l \in rank_k(\hat{\mathbf{y}})}} \frac{y_l}{p_l \log(l+1)} & \nonumber\\
nDCG@k&= \frac{DCG@k}{\sum_{l=1}^{\min(k, ||\mathbf{y}||_0)} \frac{1}{\log(l +1) }} \nonumber
 &
PSnDCG@k&= \frac{PSDCG@k}{\sum_{l=1}^{k} \frac{1}{\log l +1 }} &\nonumber,
\end{flalign}

\subsection{Further Details about Experiments and Ablation Studies}
\label{sup:xmlmeta}
\textbf{Recap of Notation:} Let us recall from Section~\ref{sec:method}, that $L$ denotes the number of labels and $V$ denotes the total number of tokens appearing across label and document texts. The training set of $N$ documents is presented as $\bc{(\vx_i,\vy_i)_{i=1}^{N}}$ with each document represented as a bag of tokens $\vx_i \in \bR^V$ with $x_{it}$ representing the TF-IDF weight of token $t \in [V]$ in the $i\nth$ document, and the ground truth label vector $\vy_i \in \bc{-1,+1}^L$ such that $y_{il} = +1$ if label $l \in [L]$ is relevant to document $i$ and $y_{il} = -1$ otherwise. For each label $l \in [L]$, its label text is similarly represented as a bag of TF-IDF scores $\vz_l \in \bR^V$. \alg learns $D$-dimensional embeddings for tokens, documents as well as labels.

\textbf{Incorporating Label text into existing BoW XML methods}: XML classifiers such as Parabel, DiSMEC, Bonsai, \etc, use a fixed BoW (bag-of-words)-based representation of documents to learn their classifiers. Label text was incorporated into these classifiers as follows: for every document $i$, let $s_{il} \in \bR$ be the relevance score the XML classifier predicted for label $l$ for document $i$. We augmented this score to incorporate label text by computing $\tilde s_{il} = \alpha \cdot s_{il} + (1-\alpha)\sigma\br{\ip{\vx_i}{\vz_l}}$. Here, $\alpha \in [0, 1]$ was fine tuned to offer the best results. Table~\ref{tab:bowmeta} shows that incorporating label text, even in this relatively crude way, still benefits accuracy.


\textbf{Generating alternative shortlists for \alg}: \alg learns a shortlister to generate a subset of labels with high recall, from an extremely large output space. Experiments were also conducted to use existing scalable XML algorithms \eg~Parabel or ANNS data structures \eg~HNSW as possible alternatives to generating this shortlist. Label centroids using learnt intermediate feature representations were provided to Parabel and HNSW in order to partition the label space. However, as Table~\ref{tab:sub:xmlclass} shows, this leads to significant reduction in precision as well as recall (upto 2\%) which adversely impacted the performance of the final ranking by \alg.

\textbf{Varying the shortlister fan-out in \alg}: \alg uses  Modules I and II to learn a shortlister. In Module I, \alg clusters the extremely large label space (in millions) to a smaller number of $K = 2^{17} \approx 130K$ meta-labels. In Module II, \alg fine-tunes the re-ranker to generate a shortlist of labels. For details of training please refer to section~\ref{sec:method} in the main paper. Experiments were conducted to observe the impact of the fan-out $K$. In particular fan-out was restricted to $2^{13} \approx 8K$ which is also a value used by contemporary algorithms such as AttentionXML and the X-Transformer. It was observed that to maintain a high recall (of around 85\%) during training \alg had to increase the beam-size by 2$\times$ which leads to increase in training time as well as a drop in accuracy (see Table~\ref{tab:combouv} \alg-8K). AttentionXML and X-Transformer were found to be computationally expensive and could not be scaled to use $2^{17}$ clusters to check whether increasing fan-out benefits them as it does \alg.


\textbf{Varying the label classifier components in \alg}: As outlined in Section~\ref{sec:method}, \alg makes crucial use of label text embeddings while learning its label classifiers $\vw_l, l \in [L]$, with two components for each label $l$ a) $\hat\vz^1_l$ that is simply the label text embedding, and b) $\hat\vz^2_l$ that is a refinement vector. $\hat\vz^2_l$ was  initialized with $\vE\vz_l$ and then fine-tuned jointly with other model parameters such as those within the residual and combination blocks, etc. An experiment was conducted in which the label embedding component $\hat\vz^1_l$ was removed from the label classifier (effectively done by setting $\hat\vz^1_l = \v0, \forall l \in [L]$) and $\hat\vz^2_l$ was randomly initialized instead. We call this configuration \alg-$\hat\vz^2$ (see Table~\ref{tab:combouv}). Another experimented was conducted to understand the importance of the refinement vector $\hat\vz^2_l$. In this experiment, $\hat\vz^2_l$ was explicitly set to $\v0$ and we used $\vw_l = \hat\vz^1_l$. We call this configuration \alg-$\hat\vz^1$ (see Table~\ref{tab:combouv}).\alg was found to be upto 5\% more accurate as compared to these variants. These experiments suggest that the novel combination of two label classifier components as proposed by \alg, namely $\hat\vz^1_l$ and $\hat\vz^2_l$ is essential for achieving high accuracy.\\

\noindent\textbf{Please go to the next page for dataset statistics and hyperparameter details.}
\begin{table*}
	\caption{Dataset Statistics. A $\ddagger$ sign denotes information that was redacted for the proprietary datasets. The first four rows are public short-text datasets. The next three rows are public full-text versions of the first three rows. The last two rows are proprietary short-text  datasets. Dataset names with an asterisk $^\ast$ next to them correspond to product-to-category tasks whereas others correspond to product-to-product tasks.}
	\label{tab:stats}
	\resizebox{\linewidth}{!}
	{
		\begin{tabular}{l|cccccccc}
			\toprule
			\textbf{Dataset} &
			\textbf{\begin{tabular}[c]{@{}c@{}}Train Documents \\ $N$ \end{tabular}} &
			\textbf{\begin{tabular}[c]{@{}c@{}}Labels\\ $L$ \end{tabular}}  &
			\textbf{\begin{tabular}[c]{@{}c@{}}Tokens\\ $V$ \end{tabular}} &
			\textbf{\begin{tabular}[c]{@{}c@{}}Test Instances \\ $N'$\end{tabular}} &
			\textbf{\begin{tabular}[c]{@{}c@{}}Average Labels\\ per Document \end{tabular}} &
			\textbf{\begin{tabular}[c]{@{}c@{}}Average Points\\ per label \end{tabular}} &
			\textbf{\begin{tabular}[c]{@{}c@{}}Average Tokens\\ per Document \end{tabular}} &
			\textbf{\begin{tabular}[c]{@{}c@{}}Average Tokens\\ per Label \end{tabular}} \\
			\midrule
			\multicolumn{9}{c}{Short text dataset statistics}\\ \midrule
            LF-AmazonTitles-131K & 294,805 & 131,073 & 40,000 & 134,835 & 2.29 & 5.15 & 7.46 & 7.15 \\
            LF-WikiSeeAlsoTitles-320K & 693,082 & 312,330 & 40,000 & 177,515 & 2.11 & 4.68 & 3.97 & 3.92 \\
            LF-WikiTitles-500K$^\ast$ & 1,813,391 & 501,070 & 80,000 & 783,743 & 4.74 & 17.15 & 3.72 & 4.16 \\
            LF-AmazonTitles-1.3M & 2,248,619 & 1,305,265 & 128,000 & 970,237 & 22.20 & 38.24 & 9.00 & 9.45 \\\midrule
            \multicolumn{9}{c}{Long text dataset statistics}\\
            \midrule
            LF-Amazon-131K & 294,805 & 131,073 & 80,000 & 134,835 & 2.29 & 5.15 & 64.28 & 4.87 \\
            LF-WikiSeeAlso-320K & 693,082 & 312,330 & 200,000 & 177,515 & 2.11 & 4.67 & 99.79 & 2.68 \\
            LF-Wikipedia-500K$^\ast$ & 1,813,391 & 501,070 & 500,000 & 783,743 & 4.74 & 17.15 & 165.18 & 3.24 \\
            \midrule
            \multicolumn{9}{c}{Proprietary dataset}\\
            \midrule
            LF-P2PTitles-300K & 1,366,429 & 300,000 & $\ddagger$ & 585,602 & $\ddagger$ & $\ddagger$ & $\ddagger$ & $\ddagger$ \\
            
            LF-P2PTitles-2M & 2,539,009 & 1,640,898 & $\ddagger$ & 1,088,146 & $\ddagger$ & $\ddagger$ & $\ddagger$ & $\ddagger$ \\
            \bottomrule
    	\end{tabular}
	}
\end{table*}
\newpage

\begin{table}
    \caption{Parameter settings for \alg on different datasets. Apart from the hyperparameters mentioned in the table below, all other hyperparameters were held constant across datasets. All ReLU layers were followed by a dropout layer with 50\% drop-rate in Module-I and 20\% for the rest of the modules. Learning rate was decayed by a decay factor of 0.5 after interval $0.5\times$ epoch length. Batch size was taken to be 255 for all datasets. Module I used 20 epochs with initial learning rate of 0.01. In Module II, 10 epochs were used with an initial learning rate of 0.008 for all datasets.}
	\label{tab:hyperparameters}
	\centering
	\begin{tabular}{l|ccc}
            \toprule
            \textbf{Dataset} &  \textbf{\begin{tabular}[c]{@{}c@{}}Beam\\ Size\end{tabular}} & \textbf{\begin{tabular}[c]{@{}c@{}}Embedding\\ Dimension\end{tabular}} & \textbf{\begin{tabular}[c]{@{}c@{}}Cluster\\ Size\end{tabular}}\\ 
            \midrule
            LF-AmazonTitles-131K &  200 & 300 & $2^{15}$\\
            LF-WikiSeeAlsoTitles-320K & 160 & 300 & $2^{17}$\\
            LF-AmazonTitles-1.3M & 100 & 512 & $2^{17}$\\
            LF-Amazon-131K &  200 & 512 & $2^{15}$\\
            LF-WikiSeeAlso-320K & 160 & 512 & $2^{17}$\\
            \midrule
            LF-P2PTitles-300K & 160 & 300 & $2^{17}$\\
            LF-P2PTitles-2M & 40 & 512 & $2^{17}$\\
            \midrule
            LF-WikiTitles-500K & 100 & 512 & $2^{17}$\\
            LF-Wikipedia-500K & 100 & 512 & $2^{17}$\\
            \bottomrule
        \end{tabular}
\end{table}

\textbf{Please go to the next page for detailed experimental results.}

\begin{table*}
\caption{A comparison of \alg on publicly available product-to-product datasets. The first 3 rows are short-text datasets whereas the last two rows are long-text versions of the first two. \alg offers predictions that are the most accurate based on all evaluation metrics, and an order of magnitude faster as compared to existing deep learning based approaches. Methods marked with a `-' sign could not be scaled for the given dataset within the available resources.}
    \label{tab:sup:prod}
    \centering
    \resizebox{\textwidth}{!}{
    \begin{tabular}{@{}c|l|ccccc|ccccc|ccc@{}}
    \toprule
    \textbf{Dataset} & \textbf{Method} & \textbf{P@1} & \textbf{P@3} & \textbf{P@5} & \textbf{N@3} & \textbf{N@5} &  \textbf{PSP@1} & \textbf{PSP@3}& \textbf{PSP@5} & \textbf{PSN@3} & \textbf{PSN@5}
    & \multicolumn{1}{c}{\begin{tabular}[c]{@{}c@{}}\textbf{Model}\\ \textbf{Size (GB)}\end{tabular}}
    & \multicolumn{1}{c}{\begin{tabular}[c]{@{}c@{}}\textbf{Training}\\ \textbf{Time (hr)}\end{tabular}}
    & \multicolumn{1}{c}{\begin{tabular}[c]{@{}c@{}}\textbf{Prediction}\\ \textbf{Time (ms)}\end{tabular}} \\
    \midrule
  \multirow{13}{*}{\textbf{\rotatebox{90}{LF-AmazonTitles-131K}}}	& \alg	 & \textbf{38.4}	 & \textbf{25.84}	 & \textbf{18.65}	 & \textbf{39.43}	 & \textbf{41.46}	 & \textbf{30.85}	 & \textbf{36.44}	 & \textbf{41.42}	 & \textbf{34.69}	 & \textbf{37.13}	 & 0.81	 & 2.16	 & 0.1\\
	& Astec	 & 37.12	 & 25.2	 & 18.24	 & 38.17	 & 40.16	 & 29.22	 & 34.64	 & 39.49	 & 32.73	 & 35.03	 & 3.24	 & 1.83	 & 2.34\\
	& AttentionXML	 & 32.25	 & 21.7	 & 15.61	 & 32.83	 & 34.42	 & 23.97	 & 28.6	 & 32.57	 & 26.88	 & 28.75	 & 2.61	 & 20.73	 & 5.19\\
	& MACH	 & 33.49	 & 22.71	 & 16.45	 & 34.36	 & 36.16	 & 24.97	 & 30.23	 & 34.72	 & 28.41	 & 30.54	 & 2.35	 & 3.3	 & 0.23\\
	& X-Transformer	 & 29.95	 & 18.73	 & 13.07	 & 28.75	 & 29.6	 & 21.72	 & 24.42	 & 27.09	 & 23.18	 & 24.39	 & -	 & -	 & 15.38\\
	& Siamese	 & 13.81	 & 8.53	 & 5.81	 & 13.32	 & 13.64	 & 13.3	 & 12.68	 & 13.36	 & 12.69	 & 13.06	 & 0.6	 & 6.92	 & 0.2\\
	& Parabel	 & 32.6	 & 21.8	 & 15.61	 & 32.96	 & 34.47	 & 23.27	 & 28.21	 & 32.14	 & 26.36	 & 28.21	 & 0.34	 & 0.03	 & 0.69\\
	& Bonsai	 & 34.11	 & 23.06	 & 16.63	 & 34.81	 & 36.57	 & 24.75	 & 30.35	 & 34.86	 & 28.32	 & 30.47	 & 0.24	 & 0.1	 & 7.49\\
	& DiSMEC	 & 35.14	 & 23.88	 & 17.24	 & 36.17	 & 38.06	 & 25.86	 & 32.11	 & 36.97	 & 30.09	 & 32.47	 & 0.11	 & 3.1	 & 5.53\\
	& PfastreXML	 & 32.56	 & 22.25	 & 16.05	 & 33.62	 & 35.26	 & 26.81	 & 30.61	 & 34.24	 & 29.02	 & 30.67	 & 3.02	 & 0.26	 & 2.32\\
	& XT	 & 31.41	 & 21.39	 & 15.48	 & 32.17	 & 33.86	 & 22.37	 & 27.51	 & 31.64	 & 25.58	 & 27.52	 & 0.84	 & 9.46	 & 9.12\\
	& Slice	 & 30.43	 & 20.5	 & 14.84	 & 31.07	 & 32.76	 & 23.08	 & 27.74	 & 31.89	 & 26.11	 & 28.13	 & 0.39	 & 0.08	 & 1.58\\
	& AnneXML	 & 30.05	 & 21.25	 & 16.02	 & 31.58	 & 34.05	 & 19.23	 & 26.09	 & 32.26	 & 23.64	 & 26.6	 & 1.95	 & 0.08	 & 0.11\\
	 \midrule
	 \multirow{13}{*}{\textbf{\rotatebox{90}{LF-WikiSeeAlsoTitles-320K}}}	& \alg	 & \textbf{25.14}	 & \textbf{16.9}	 & \textbf{12.86}	 & \textbf{24.99}	 & \textbf{25.95}	 & \textbf{16.73}	 & \textbf{18.99}	 & \textbf{21.01}	 & \textbf{19.18}	 & \textbf{20.75}	 & 1.76	 & 11.16	 & 0.09\\
	& Astec	 & 22.72	 & 15.12	 & 11.43	 & 22.16	 & 22.87	 & 13.69	 & 15.81	 & 17.5	 & 15.56	 & 16.75	 & 7.3	 & 4.17	 & 2.67\\
	& AttentionXML	 & 17.56	 & 11.34	 & 8.52	 & 16.58	 & 17.07	 & 9.45	 & 10.63	 & 11.73	 & 10.45	 & 11.24	 & 6.02	 & 56.12	 & 7.08\\
	& MACH	 & 18.06	 & 11.91	 & 8.99	 & 17.57	 & 18.17	 & 9.68	 & 11.28	 & 12.53	 & 11.19	 & 12.14	 & 2.51	 & 8.23	 & 0.52\\
	& X-Transformer	 & -	 & -	 & -	 & -	 & -	 & -	 & -	 & -	 & -	 & -	 & -	 & -	 & -\\
	& Siamese	 & 10.69	 & 6.28	 & 4.51	 & 9.79	 & 9.91	 & 10.1	 & 9.43	 & 9.59	 & 10.22	 & 10.47	 & 0.67	 & 11.58	 & 0.17\\
	& Parabel	 & 17.68	 & 11.48	 & 8.59	 & 16.96	 & 17.44	 & 9.24	 & 10.65	 & 11.8	 & 10.49	 & 11.32	 & 0.6	 & 0.07	 & 0.8\\
	& Bonsai	 & 19.31	 & 12.71	 & 9.55	 & 18.74	 & 19.32	 & 10.69	 & 12.44	 & 13.79	 & 12.29	 & 13.29	 & 0.37	 & 0.37	 & 14.82\\
	& DiSMEC	 & 19.12	 & 12.93	 & 9.87	 & 18.93	 & 19.71	 & 10.56	 & 13.01	 & 14.82	 & 12.7	 & 14.02	 & 0.19	 & 15.56	 & 11.02\\
	& PfastreXML	 & 17.1	 & 11.13	 & 8.35	 & 16.8	 & 17.35	 & 12.15	 & 12.51	 & 13.26	 & 12.81	 & 13.48	 & 6.77	 & 0.59	 & 2.59\\
	& XT	 & 17.04	 & 11.31	 & 8.6	 & 16.61	 & 17.24	 & 8.99	 & 10.52	 & 11.82	 & 10.33	 & 11.26	 & -	 & 5.28	 & 12.86\\
	& Slice	 & 18.55	 & 12.62	 & 9.68	 & 18.29	 & 19.07	 & 11.24	 & 13.45	 & 15.2	 & 13.03	 & 14.23	 & 0.94	 & 0.2	 & 1.85\\
	& AnneXML	 & 16.3	 & 11.24	 & 8.84	 & 16.19	 & 17.14	 & 7.24	 & 9.63	 & 11.75	 & 9.06	 & 10.43	 & 4.22	 & 0.21	 & 0.13\\
	 \midrule
	\multirow{11}{*}{\textbf{\rotatebox{90}{LF-AmazonTitles-1.3M}}}	& \alg	 & \textbf{50.67}	 & \textbf{44.49}	 & \textbf{40.35}	 & \textbf{48.05}	 & \textbf{46.85}	 & 22.07	 & 26.54	 & 29.3	 & 25.06	 & 26.85	 & 9.62	 & 74.47	 & 0.16\\
	& Astec	 & 48.82	 & 42.62	 & 38.44	 & 46.11	 & 44.8	 & 21.47	 & 25.41	 & 27.86	 & 24.08	 & 25.66	 & 26.66	 & 18.54	 & 2.61\\
	& AttentionXML	 & 45.04	 & 39.71	 & 36.25	 & 42.42	 & 41.23	 & 15.97	 & 19.9	 & 22.54	 & 18.23	 & 19.6	 & 28.84	 & 380.02	 & 29.53\\
	& MACH	 & 35.68	 & 31.22	 & 28.35	 & 33.42	 & 32.27	 & 9.32	 & 11.65	 & 13.26	 & 10.79	 & 11.65	 & 7.68	 & 60.39	 & 2.09\\
	& X-Transformer	 & -	 & -	 & -	 & -	 & -	 & -	 & -	 & -	 & -	 & -	 & -	 & -	 & -\\
	& Siamese	 & -	 & -	 & -	 & -	 & -	 & -	 & -	 & -	 & -	 & -	 & -	 & -	 & -\\
	& Parabel	 & 46.79	 & 41.36	 & 37.65	 & 44.39	 & 43.25	 & 16.94	 & 21.31	 & 24.13	 & 19.7	 & 21.34	 & 11.75	 & 1.5	 & 0.89\\
	& Bonsai	 & 47.87	 & 42.19	 & 38.34	 & 45.47	 & 44.35	 & 18.48	 & 23.06	 & 25.95	 & 21.52	 & 23.33	 & 9.02	 & 7.89	 & 39.03\\
	& DiSMEC	 & -	 & -	 & -	 & -	 & -	 & -	 & -	 & -	 & -	 & -	 & -	 & -	 & -\\
	& PfastreXML	 & 37.08	 & 33.77	 & 31.43	 & 36.61	 & 36.61	 & \textbf{28.71}	 & \textbf{30.98}	 & \textbf{32.51}	 & \textbf{29.92}	 & \textbf{30.73}	 & 29.59	 & 9.55	 & 23.64\\
	& XT	 & 40.6	 & 35.74	 & 32.01	 & 38.18	 & 36.68	 & 13.67	 & 17.11	 & 19.06	 & 15.64	 & 16.65	 & 7.9	 & 82.18	 & 5.94\\
	& Slice	 & 34.8	 & 30.58	 & 27.71	 & 32.72	 & 31.69	 & 13.8	 & 16.87	 & 18.89	 & 15.62	 & 16.74	 & 5.98	 & 0.79	 & 1.45\\
	& AnneXML	 & 47.79	 & 41.65	 & 36.91	 & 44.83	 & 42.93	 & 15.42	 & 19.67	 & 21.91	 & 18.05	 & 19.36	 & 14.53	 & 2.48	 & 0.12\\
	\midrule
	\multirow{11}{*}{\textbf{\rotatebox{90}{LF-Amazon-131K}}}	& \alg	 & \textbf{42.94}	 & 28.79	 & \textbf{21}	 & \textbf{44.25}	 & \textbf{46.84}	 & \textbf{34.52}	 & \textbf{41.14}	 & \textbf{47.33}	 & \textbf{39.35}	 & \textbf{42.48}	 & 1.86	 & 1.8	 & 0.1\\
	& Astec	 & -	 & -	 & -	 & -	 & -	 & -	 & -	 & -	 & -	 & -	 & -	 & -	 & -\\
	& AttentionXML	 & 42.9	 & \textbf{28.96}	 & 20.97	 & 44.07	 & 46.44	 & 32.92	 & 39.51	 & 45.24	 & 37.49	 & 40.33	 & 5.04	 & 50.17	 & 12.33\\
	& MACH	 & 34.52	 & 23.39	 & 17	 & 35.53	 & 37.51	 & 25.27	 & 30.71	 & 35.42	 & 29.02	 & 31.33	 & 4.57	 & 13.91	 & 0.25\\
	& X-Transformer	 & -	 & -	 & -	 & -	 & -	 & -	 & -	 & -	 & -	 & -	 & -	 & -	 & -\\
	& Bonsai	 & 40.23	 & 27.29	 & 19.87	 & 41.46	 & 43.84	 & 29.6	 & 36.52	 & 42.39	 & 34.43	 & 37.34	 & 0.46	 & 0.4	 & 7.41\\
	& DiSMEC	 & 41.68	 & 28.32	 & 20.58	 & 43.22	 & 45.69	 & 31.61	 & 38.96	 & 45.07	 & 36.97	 & 40.05	 & 0.45	 & 7.12	 & 15.48\\
	& PfastreXML	 & 35.83	 & 24.35	 & 17.6	 & 36.97	 & 38.85	 & 28.99	 & 33.24	 & 37.4	 & 31.65	 & 33.62	 & 0.01	 & 1.54	 & 3.32\\
	& XT	 & 34.31	 & 23.27	 & 16.99	 & 35.18	 & 37.26	 & 24.35	 & 29.81	 & 34.7	 & 27.95	 & 30.34	 & 0.92	 & 1.38	 & 7.42\\
	& Slice	 & 32.07	 & 22.21	 & 16.52	 & 33.54	 & 35.98	 & 23.14	 & 29.08	 & 34.63	 & 27.25	 & 30.06	 & 0.39	 & 0.11	 & 1.35\\
	& AnneXML	 & 35.73	 & 25.46	 & 19.41	 & 37.81	 & 41.08	 & 23.56	 & 31.97	 & 39.95	 & 29.07	 & 33	 & 4.01	 & 0.68	 & 0.11\\
	 \midrule
	 \multirow{11}{*}{\textbf{\rotatebox{90}{LF-WikiSeeAlso-320K}}}	& \alg	 & \textbf{41.36}	 & \textbf{28.04}	 & \textbf{21.38}	 & \textbf{41.55}	 & \textbf{43.32}	 & \textbf{25.72}	 & \textbf{30.93}	 & \textbf{34.89}	 & \textbf{30.69}	 & \textbf{33.69}	 & 4.84	 & 13.4	 & 0.09\\
	& Astec	 & -	 & -	 & -	 & -	 & -	 & -	 & -	 & -	 & -	 & -	 & -	 & -	 & -\\
	& AttentionXML	 & 40.5	 & 26.43	 & 19.87	 & 39.13	 & 40.26	 & 22.67	 & 26.66	 & 29.83	 & 26.13	 & 28.38	 & 7.12	 & 90.37	 & 12.6\\
	& MACH	 & 27.18	 & 17.38	 & 12.89	 & 26.09	 & 26.8	 & 13.11	 & 15.28	 & 16.93	 & 15.17	 & 16.48	 & 11.41	 & 50.22	 & 0.54\\
	& X-Transformer	 & -	 & -	 & -	 & -	 & -	 & -	 & -	 & -	 & -	 & -	 & -	 & -	 & -\\
	& Bonsai	 & 34.86	 & 23.21	 & 17.66	 & 34.09	 & 35.32	 & 18.19	 & 22.35	 & 25.66	 & 21.62	 & 23.84	 & 0.84	 & 1.39	 & 8.94\\
	& DiSMEC	 & 34.59	 & 23.58	 & 18.26	 & 34.43	 & 36.11	 & 18.95	 & 23.92	 & 27.9	 & 23.04	 & 25.76	 & 1.28	 & 58.79	 & 75.52\\
	& PfastreXML	 & 28.79	 & 18.38	 & 13.6	 & 27.69	 & 28.28	 & 17.12	 & 18.19	 & 19.43	 & 18.23	 & 19.2	 & 14.02	 & 4.97	 & 2.68\\
	& XT	 & 30.1	 & 19.6	 & 14.92	 & 28.65	 & 29.58	 & 14.43	 & 17.13	 & 19.69	 & 16.37	 & 17.97	 & 2.2	 & 3.27	 & 4.79\\
	& Slice	 & 27.74	 & 19.39	 & 15.47	 & 27.84	 & 29.65	 & 13.07	 & 17.5	 & 21.55	 & 16.36	 & 18.9	 & 0.94	 & 0.2	 & 1.18\\
	& AnneXML	 & 30.79	 & 20.88	 & 16.47	 & 30.02	 & 31.64	 & 13.48	 & 17.92	 & 22.21	 & 16.52	 & 19.08	 & 12.13	 & 2.4	 & 0.11\\
    \bottomrule
    \end{tabular}
    }
\end{table*}

\begin{table*}
\caption{A comparison of \alg's performance on product-to-category datasets. The first row is a short-text dataset and the second row its long-text counterpart. Although \alg focuses on product-to-product tasks, it is nevertheless competitive in terms of accuracy, as well as an order of magnitude faster in prediction as compared to leading deep learning approaches. Methods marked with a `-' sign could not be scaled for the given dataset within the available resources. The AttentionXML method used a non-standard version of the Wikipedia-500K dataset. All other methods, including \alg, used the standard version of the dataset.}
    \label{tab:sup:category}
    \centering
    \resizebox{\textwidth}{!}{
    \begin{tabular}{@{}c|l|ccccc|ccccc|ccc@{}}
    \toprule
    \textbf{Dataset} & \textbf{Method} & \textbf{P@1} & \textbf{P@3} & \textbf{P@5} & \textbf{N@3} & \textbf{N@5} &  \textbf{PSP@1} & \textbf{PSP@3}& \textbf{PSP@5} & \textbf{PSN@3} & \textbf{PSN@5}
    & \multicolumn{1}{c}{\begin{tabular}[c]{@{}c@{}}\textbf{Model}\\ \textbf{Size (GB)}\end{tabular}}
    & \multicolumn{1}{c}{\begin{tabular}[c]{@{}c@{}}\textbf{Training}\\ \textbf{Time (hrs)}\end{tabular}}
    & \multicolumn{1}{c}{\begin{tabular}[c]{@{}c@{}}\textbf{Prediction}\\ \textbf{Time (ms)}\end{tabular}} \\
    \midrule
  
	\multirow{11}{*}{\textbf{\rotatebox{90}{LF-WikiTitles-500K}}}	& \alg	 & 44.21	 & 24.64	 & 17.36	 & \textbf{33.55}	 & \textbf{31.92}	 & \textbf{19.29}	 & \textbf{19.82}	 & \textbf{19.96}	 & \textbf{21.26}	 & \textbf{22.95}	 & 4.53	 & 42.26	 & 0.09\\
	& Astec-3	 & \textbf{44.4}	 & \textbf{24.69}	 & \textbf{17.49}	 & 33.43	 & 31.72	 & 18.31	 & 18.25	 & 18.56	 & 19.57	 & 21.09	 & 15.01	 & 13.5	 & 2.7\\
	& AttentionXML	 & 40.9	 & 21.55	 & 15.05	 & 29.38	 & 27.45	 & 14.8	 & 13.97	 & 13.88	 & 15.24	 & 16.22	 & 14.01	 & 133.94	 & 9\\
	& MACH	 & 37.74	 & 19.11	 & 13.26	 & 26.63	 & 24.94	 & 13.71	 & 12.14	 & 12	 & 13.63	 & 14.54	 & 4.73	 & 22.46	 & 0.8\\
	& X-Transformer	 & -	 & -	 & -	 & -	 & -	 & -	 & -	 & -	 & -	 & -	 & -	 & -	 & -\\
	& Siamese	 & -	 & -	 & -	 & -	 & -	 & -	 & -	 & -	 & -	 & -	 & -	 & -	 & -\\
	& Parabel	 & 40.41	 & 21.98	 & 15.42	 & 29.89	 & 28.15	 & 15.55	 & 15.32	 & 15.35	 & 16.5	 & 17.66	 & 2.7	 & 0.42	 & 0.81\\
	& Bonsai	 & 40.97	 & 22.3	 & 15.66	 & 30.35	 & 28.65	 & 16.58	 & 16.34	 & 16.4	 & 17.6	 & 18.85	 & 1.63	 & 2.03	 & 17.38\\
	& DiSMEC	 & 39.42	 & 21.1	 & 14.85	 & 28.87	 & 27.29	 & 15.88	 & 15.54	 & 15.89	 & 16.76	 & 18.13	 & 0.68	 & 48.27	 & 11.71\\
	& PfastreXML	 & 35.71	 & 19.27	 & 13.64	 & 26.45	 & 25.15	 & 18.23	 & 15.42	 & 15.08	 & 17.34	 & 18.24	 & 20.41	 & 3.79	 & 9.37\\
	& XT	 & 38.19	 & 20.74	 & 14.68	 & 28.15	 & 26.64	 & 14.2	 & 14.14	 & 14.41	 & 15.18	 & 16.45	 & 3.1	 & 8.78	 & 7.56\\
	& Slice	 & 25.48	 & 15.06	 & 10.98	 & 20.67	 & 20.52	 & 13.9	 & 13.33	 & 13.82	 & 14.5	 & 15.9	 & 2.3	 & 0.74	 & 1.76\\
	& AnneXML	 & 39	 & 20.66	 & 14.55	 & 28.4	 & 26.8	 & 13.91	 & 13.38	 & 13.75	 & 14.63	 & 15.88	 & 11.18	 & 1.98	 & 0.13\\
	\midrule
	\multirow{11}{*}{\textbf{\rotatebox{90}{LF-Wikipedia-500K}}}	& \alg	 & 73.96	 & 54.17	 & 42.43	 & 66.31	 & 64.81	 & 32.13	 & 40.13	 & 44.59	 & 39.57	 & 43.7	 & 9.34	 & 44.23	 & 0.09\\
	& Astec	 & -	 & -	 & -	 & -	 & -	 & -	 & -	 & -	 & -	 & -	 & -	 & -	 & -\\
	& AttentionXML	 & \textbf{82.73}	 & \textbf{63.75}	 & \textbf{50.41}	 & \textbf{76.56}	 & \textbf{74.86}	 & \textbf{34}	 & \textbf{44.32}	 & \textbf{50.15}	 & \textbf{42.99}	 & \textbf{47.69}	 & 9.73	 & 221.6	 & 12.38\\
	& MACH	 & 52.48	 & 31.93	 & 23.34	 & 41.7	 & 39.43	 & 17.92	 & 18.16	 & 18.66	 & 19.45	 & 20.77	 & 28.12	 & 220.07	 & 0.82\\
	& X-Transformer	 & -	 & -	 & -	 & -	 & -	 & -	 & -	 & -	 & -	 & -	 & -	 & -	 & -\\
	& Siamese	 & -	 & -	 & -	 & -	 & -	 & -	 & -	 & -	 & -	 & -	 & -	 & 0.03	 & -\\
	& Parabel	 & 70.14	 & 50.62	 & 39.45	 & 61.86	 & 59.89	 & 27.25	 & 32.52	 & 35.93	 & 32.29	 & 35.31	 & 5.51	 & 3.02	 & 2.01\\
	& Bonsai	 & 70.56	 & 51.11	 & 39.86	 & 62.47	 & 60.61	 & 28.18	 & 33.86	 & 37.55	 & 33.58	 & 36.86	 & 3.94	 & 17.22	 & 22.23\\
	& DiSMEC	 & -	 & -	 & -	 & -	 & -	 & -	 & -	 & -	 & -	 & -	 & -	 & -	 & -\\
	& PfastreXML	 & 61.24	 & 41.59	 & 31.75	 & 52.26	 & 50.34	 & 33.3	 & 32.56	 & 33.67	 & 33.77	 & 35.25	 & 48.26	 & 24.71	 & 7.69\\
	& XT	 & 66.98	 & 48.33	 & 37.82	 & 58.94	 & 57.19	 & 24.78	 & 30.06	 & 33.46	 & 29.63	 & 32.51	 & 3.9	 & 16.73	 & 3.81\\
	& Slice	 & 47.51	 & 32.34	 & 25.07	 & 40.56	 & 39.51	 & 19.6	 & 21.99	 & 24.6	 & 22.2	 & 24.53	 & 2.3	 & 0.67	 & 1.58\\
	& AnneXML	 & 64.77	 & 43.24	 & 32.79	 & 54.63	 & 52.51	 & 24.08	 & 28.25	 & 31.2	 & 28.47	 & 31.3	 & 49.25	 & 14.97	 & 5.15\\
	 \bottomrule
    \end{tabular}
    }
\end{table*}

\begin{table*}
\caption{A comparison of \alg's performance on the proprietary datasets. \alg can be an order of magnitude faster in prediction as compared to existing deep learning approaches.}
    \label{tab:sup:p2p}
    \centering
    \resizebox{\textwidth}{!}{
    \begin{tabular}{@{}c|l|ccccc|ccccc@{}}
    \toprule
    \textbf{Dataset} & \textbf{Method} & \textbf{P@1} & \textbf{P@3} & \textbf{P@5} & \textbf{N@3} & \textbf{N@5} &  \textbf{PSP@1} & \textbf{PSP@3}& \textbf{PSP@5} & \textbf{PSN@3} & \textbf{PSN@5}\\
    \midrule
    \multirow{5}{*}{\textbf{\rotatebox{45}{P2PTitles-300K}}}	& \alg	 & \textbf{47.17}	 & \textbf{30.67}	 & \textbf{22.69}	 & \textbf{53.62}	 & \textbf{57.06}	 & \textbf{42.43}	 & \textbf{55.07}	 & \textbf{62.3}	 & \textbf{49.86} &  \textbf{53.27}\\
	& Astec	 & 44.3	 & 28.95	 & 21.56	 & 50.36	 & 53.67	 & 39.44	 & 50.9	 & 57.83	 & 45.99	 & 49.12	 \\
	& Parabel	 & 43.14	 & 28.34	 & 20.99	 & 48.73	 & 51.75	 & 37.26	 & 48.87	 & 55.32	 & 43.45	 & 46.32\\
	& PfastreXML	 & 39.4	 & 25.6	 & 18.77	 & 44.59	 & 46.98	 & 35.79	 & 45.13	 & 49.9	 & 40.98	 & 43.03	 \\
	& Slice	 & 31.27	 & 28.91	 & \textbf{25.19}	 & 31.5	 & 33.2	 & 27.03	 & 30.44	 & 34.95	 & 28.54	 & 30.77 \\
	\midrule
  \multirow{5}{*}{\textbf{\rotatebox{45}{P2PTitles-2M}}}	& \alg	 & \textbf{40.27}	 & \textbf{36.65}	 & \textbf{31.45}	 & \textbf{40.4}	 & \textbf{42.49}	 & \textbf{36.65}	 & \textbf{40.14}	 & \textbf{45.15}	 & \textbf{38.23}	 & \textbf{40.99}	\\
	& Astec	 & 36.34	 & 33.33	 & 28.74	 & 36.63	 & 38.63	 & 32.75	 & 36.3	 & 41	 & 34.43	 & 36.97\\
	& Parabel	 & 35.26	 & 32.44	 & 28.06	 & 35.3	 & 36.89	 & 30.21	 & 33.85	 & 38.46	 & 31.63	 & 33.71\\
	& PfastreXML	 & 30.52	 & 28.68	 & 24.6	 & 31.5	 & 33.23	 & 28.84	 & 32.1	 & 35.65	 & 30.56	 & 32.52	 \\
	& Slice	 & 31.27	 & 28.91	 & 25.19	 & 31.5	 & 33.2	 & 27.03	 & 30.44	 & 34.95	 & 28.54	 & 30.77\\
	 \bottomrule
    \end{tabular}
    }
\end{table*}

\clearpage

\begin{longtable}{c|p{0.75\linewidth}}
 \caption{A subjective comparison of \alg's prediction quality as compared to state-of-the-art deep learning, as well as BoW approaches on examples taken from the test sets of two datasets. Predictions in black color in non-bold/non-italic font were a part of the ground truth. Predictions in bold italics were such that their co-occurrence with other ground truth labels in the test set was novel, i.e. those co-occurences were never seen in the training set. Predictions in light gray color were not a part of the ground truth. \alg offers much more precise recommendations on these examples as compared to other methods, for example AttentionXML, whose predictions on the last example are mostly irrelevant, e.g. focusing on labels such as ``Early United States commemorative coins'', instead of those related to the New Zealand dollar. \alg is able to predict labels that never co-occur in the training set due to its inclusion of label text in to the classifier. For example, in the last example, the label ``Australian dollar'' never occurred with the other ground truth labels in the training set i.e. had no common training instances with the rest of the ground truth labels. Similarly, in the first example, the label ``Panzer Dragoon Orta'' never occurred together with other ground truth labels yet \alg predicted these labels correctly while the other XML algorithms could not do so. \label{tab:examples}} \\
        \toprule
        \textbf{Algorithm} & \textbf{Predictions} \\
         \midrule
         \midrule
        \multicolumn{2}{c}{LF-AmazonTitles-131K} \\
        \midrule
        \midrule
         \textbf{Document} & \textbf{Panzer Dragoon Zwei} \\
         \midrule
         \alg &  Panzer Dragoon, Action Replay Plus, Sega Saturn System - Video Game Console, \wpred{The Legend of Dragoon}, \emph{\textbf{Panzer Dragoon Orta}} \\
         \midrule
         Astec & \wpred{Guns of the Wehrmacht 1933-1945 (2006), Mission Barbarossa, Stug III \& IV-Assault Guns, Tiger: Heavy Tank Panzer VI, Blitzkrieg} \\
         \midrule
         Bonsai & \wpred{Playstation 1 Memory Card (1 MB),   Mission Barbarossa, PlayStation 2 Memory Card (8MB), The Legend of Dragoon, Blitzkrieg} \\
         \midrule
         MACH & \wpred{Mission Barbarossa, German Military Vehicles, Guns of the Wehrmacht 1933-1945 (2006), Stug III \& IV - Assault Guns, Panther - The Panzer V (2006)}\\
         \midrule
         AttentionXML & \wpred{Panther - The Panzer V (2006), Mission Barbarossa, German Military Vehicles, Stug III \& IV - Assault Guns, The Legend of Zelda: A Link to the Past} \\
         \midrule
         Slice & \wpred{Tiger: Heavy Tank Panzer VI, Stug III \& IV - Assault Guns, Guns of the Wehrmacht 1933-1945 (2006), German Military Vehicles, The Legend of Dragoon}\\
        \midrule
        \midrule
        \textbf{Document} & \textbf{Wagner - Die Walkure / Gambill, Denoke, Rootering, Behle, Jun, Vaughn, Zagrosek, Stuttgart Opera} \\
        \midrule
        DECAF & Wagner - Siegfried / West, Gasteen, Göhring, Schöne, Waag, Jun, Herrera, Zagrosek, Stuttgart Opera, Wagner - Gotterdammerung / Bonnema, DeVol, Iturralde, Kapellmann, Bracht, Westbroek, Zagrosek, Stuttgart Opera, Wagner - Der Fliegende Holländer (1986), \emph{\textbf{Wagner - Gotterdammerung / Treleaven, Polaski, Salminen, Struckmann, Matos, von Kannen, de Billy, Barcelona Opera (2005)}}, \wpred{Seligpreisung} \\
        \midrule
        Astec & \wpred{Lehar: Die lustige Witwe (2004)}, Wagner - Der Fliegende Holländer (1986), \wpred{Wagner: Der Ring des Nibelungen - Complete Ring Cycle (Levine, Metropolitan Opera) (2002), Verdi: Macbeth (The Metropolitan Opera HD Live Series) (2008), Richard Wagner - Der fliegende Holländer (2008)} \\
        \midrule
        Bonsai & \wpred{Puccini: Tosca (Royal Opera House 2011) (2012), Weisser Holunder, Wagner - Die Meistersinger von Nurnberg / Heppner, Mattila, Morris, Pape, Allen, Polenzani, Levine, Metropolitan Opera, Wagner: Parsifal / Metropolitan Opera (1993), Un Giorno Di Regno (2010)} \\
        \midrule
        MACH & \wpred{Puccini: Tosca (Royal Opera House 2011) (2012), Puccini: La Boheme, The Main Event [ORIGINAL RECORDINGS REMASTERED] 2CD SET, Super Mario World}, Wagner - Gotterdammerung / Bonnema, DeVol, Iturralde, Kapellmann, Bracht, Westbroek, Zagrosek, Stuttgart Opera \\
        \midrule
        Parabel & Wagner - Der Fliegende Holländer (1986), \wpred{Die Meistersinger Von Nurnberg [Blu-ray] (2011), Un Giorno Di Regno (2010), Puccini: La Boheme, Wagner - Die Meistersinger von Nurnberg / Heppner, Mattila, Morris, Pape, Allen, Polenzani, Levine, Metropolitan Opera} \\
        \midrule
        AttentionXML & \wpred{Puccini: La Boheme, Verdi: Macbeth, Puccini: Tosca (Royal Opera House 2011) (2012), Tannhauser (2008)}, Wagner - Gotterdammerung / Treleaven, Polaski, Salminen, Struckmann, Matos, von Kannen, de Billy, Barcelona Opera (2005) \\
        \midrule
        Slice & \wpred{Rossini - Semiramide / Conlon, Anderson, Horne, Metropolitan Opera (1991), Richard Wagner - Der fliegende Holländer (2008)}, Wagner - Der Fliegende Holländer (1986), \wpred{Wagner - Gotterdammerung / Jones, Mazura, Jung, Hubner, Becht, Altmeyer, Killebrew, Boulez, Bayreuth Opera (Boulez Ring Cycle Part 4) (2005), Rossini - Il Turco in Italia / Bartoli, Raimondi, Macias, Rumetz, Schmid, Welser-Most, Zurich Opera (2004)} \\
        \midrule
        \midrule
        \multicolumn{2}{c}{LF-WikiSeeAlsoTitles-320K} \\
        \midrule
        \midrule
        \textbf{Document} & \textbf{New Zealand dollar} \\
        \midrule
        DECAF & \emph{\textbf{Economy of New Zealand}}, Cook Islands dollar, \wpred{Politics of New Zealand}, Pitcairn Islands dollar, \emph{\textbf{Australian dollar}} \\
        \midrule
        Astec & \wpred{Coins of the Australian dollar, List of banks in New Zealand, Constitution of New Zealand, Independence of New Zealand, History of New Zealand} \\
        \midrule
        Bonsai & \wpred{Military history of New Zealand, History of New Zealand, Timeline of New Zealand history, Timeline of the New Zealand environment, List of years in New Zealand} \\
        \midrule
        MACH & \wpred{Military history of New Zealand, History of New Zealand, List of years in New Zealand, Timeline of the New Zealand environment, Timeline of New Zealand's links with Antarctica} \\
        \midrule
        Parabel & \wpred{Early United States commemorative coins, Environment of New Zealand, Half dollar (United States coin), History of New Zealand, Conservation in New Zealand} \\
        \midrule
        AttentionXML & \wpred{Coins of the Australian dollar, Early United States commemorative coins, Half dollar (United States coin), Agriculture in New Zealand, Politics of New Zealand} \\
        \midrule
        Slice & \wpred{Timeline of New Zealand's links with Antarctica, Coins of the Australian dollar, Early United States commemorative coins, List of New Zealand state highways, Timeline of the New Zealand environment} \\
         \bottomrule
\end{longtable}


\end{document}